\documentclass{article}

\usepackage{graphicx}
\usepackage{wrapfig}

\usepackage{subcaption}
\usepackage{makecell}
\usepackage{tabulary}
\usepackage{multirow}
\usepackage[margin=1in]{geometry}
\usepackage{titlesec}
\usepackage[labelfont=bf,font=footnotesize]{caption}
\usepackage{parskip}
\usepackage{mathpazo}
\usepackage[comma,sort&compress,numbers]{natbib}
\usepackage{booktabs}
\usepackage{listings}
\usepackage{float}

\usepackage{amssymb}
\usepackage{amsthm}
\usepackage{amsmath}
\usepackage{nicefrac}        
\usepackage{siunitx}

\graphicspath{{Figures/}}
\DeclareGraphicsExtensions{.pdf,.png,.jpg}

\titlespacing\section{0pt}{6pt plus 1pt minus 0pt}{3pt plus 1pt minus 0pt}
\titlespacing\subsection{0pt}{4pt plus 1pt minus 0pt}{3pt plus 1pt minus 0pt}
\titlespacing\subsubsection{0pt}{4pt plus 1pt minus 1pt}{3pt plus 1pt minus 1pt}
\titleformat{\section}{\large\bfseries\sffamily}{\thesection}{1em}{}
\titleformat{\subsection}{\normalsize\bfseries\sffamily}{\thesubsection}{1em}{}
\titleformat{\subsubsection}{\normalsize\bfseries\sffamily}{\thesubsubsection}{1em}{}

\usepackage[ruled]{algorithm2e}

\usepackage{flushend}

\usepackage{tabulary}
\usepackage{multirow}

\usepackage{array}
\usepackage{enumitem}
\usepackage{todonotes}
\usepackage{url}
\usepackage[T1]{fontenc}

\usepackage{hyperref}
\hypersetup{
    colorlinks,
    linkcolor={blue},
    citecolor={blue},
    urlcolor={blue!80!black},
    breaklinks=true,
	plainpages=true
}
\newcommand{\cref}[3]{\hyperref[#2]{#1~\ref*{#2}{#3}}}
\newcommand{\colref}[2]{\hyperref[#2]{#1~\ref*{#2}}}

\newcommand{\figref}[1]{\colref{Fig.}{#1}}

\newcommand{\secref}[1]{\colref{Section}{#1}}

\newcommand{\tabref}[1]{\colref{Table}{#1}}
\newcommand{\coloredref}[2]{\hyperref[#2]{#1~\ref*{#2}}}
\newcommand{\coloredsubref}[3]{\hyperref[#2]{#1~\ref*{#2}{#3}}}
\newcommand{\comment}[1]{}

\begin{document}

\begin{center}
{\usefont{OT1}{phv}{b}{n}\selectfont\Large{ProFusion: 3D Reconstruction of Protein Complex Structures from Multi-view AFM Images}}

{\usefont{OT1}{phv}{}{}\selectfont\small
{Jaydeep Rade$^{1,+}$, Md Hasibul Hasan Hasib$^{1,+}$, Meric Ozturk$^3$, Baboucarr Faal$^5$, Sheng Yang$^{3,4}$, Dipali G. Sashital$^3$, Vincenzo Venditti$^5$, Baoyu Chen$^{3,6}$, Soumik Sarkar$^2$, Adarsh Krishnamurthy$^{2*}$, Anwesha Sarkar$^{1*}$}}

{\usefont{OT1}{phv}{}{}\selectfont\scriptsize
{$^1$ Department of Electrical and Computer Engineering, Iowa State University, Ames, Iowa 50011, USA\\
$^2$ Department of Mechanical Engineering, Iowa State University, Ames, Iowa 50011, USA\\
$^3$ Roy J. Carver Department of Biochemistry, Biophysics and Molecular Biology, Iowa State University, Ames, Iowa 50011, USA\\
$^4$ Target \& Protein Sciences, Janssen R\&D, Johnson \& Johnson, 1400 McKean Rd, Spring house, Pennsylvania 19477, USA.\\
$^5$ Department of Chemistry, Iowa State University, Ames, Iowa 50011, USA\\
$^6$ Departments of Internal Medicine and Biophysics, University of Texas Southwestern Medical Center, Dallas, Texas 75390, USA\\
* Corresponding authors: \url{adarsh@iastate.edu}, \url{anweshas@iastate.edu}
}}
\end{center}

\section*{Abstract}
AI-based \emph{in silico} methods have improved protein structure prediction but often struggle with large protein complexes (PCs) involving multiple interacting proteins due to missing 3D spatial cues. Experimental techniques like Cryo-EM are accurate but costly and time-consuming. We present ProFusion, a hybrid framework that integrates a deep learning model with Atomic Force Microscopy (AFM), which provides high-resolution height maps from random orientations, naturally yielding multi-view data for 3D reconstruction. However, generating a large-scale AFM imaging data set sufficient to train deep learning models is impractical. Therefore, we developed a \textit{virtual} AFM framework that simulates the imaging process and generated a dataset of ~542,000 proteins with multi-view synthetic AFM images. We train a conditional diffusion model to synthesize novel views from unposed inputs and an instance-specific Neural Radiance Field (NeRF) model to reconstruct 3D structures. Our reconstructed 3D protein structures achieve an average Chamfer Distance within the AFM imaging resolution, reflecting high structural fidelity. Our method is extensively validated on experimental AFM images of various PCs, demonstrating strong potential for accurate, cost-effective protein complex structure prediction and rapid iterative validation using AFM experiments.

\subsection*{Keywords}
Protein Structure Prediction~$|$ Atomic Force Microscopy~$|$ Multi-view 3D Reconstruction~$|$ Diffusion Models~$|$ Neural Radiance Fields (NeRFs)

\section{Introduction}\label{intro}
\begin{figure}[h!]
    \centering
    \begin{subfigure}[c]{\linewidth}
        \centering
        \includegraphics[width=0.9\linewidth, trim={0in 2in 0in 1.75in},clip]{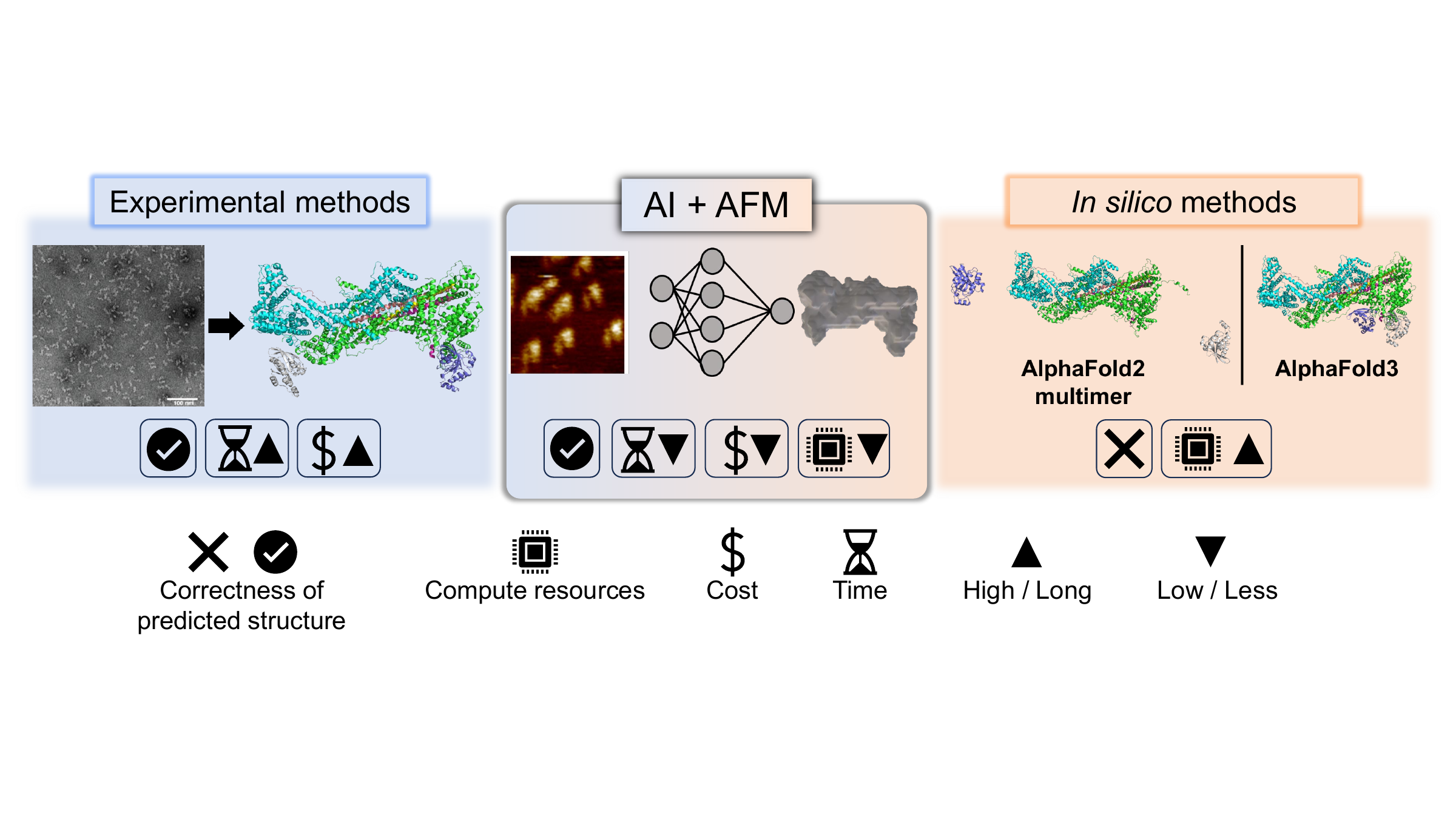}
        \caption{}
        \label{fig:motivation_overview}
    \end{subfigure}
    \begin{subfigure}[c]{\linewidth}
        \centering
        \includegraphics[width=0.9\linewidth, trim={0in 2.1in 0in 1.75in},clip]{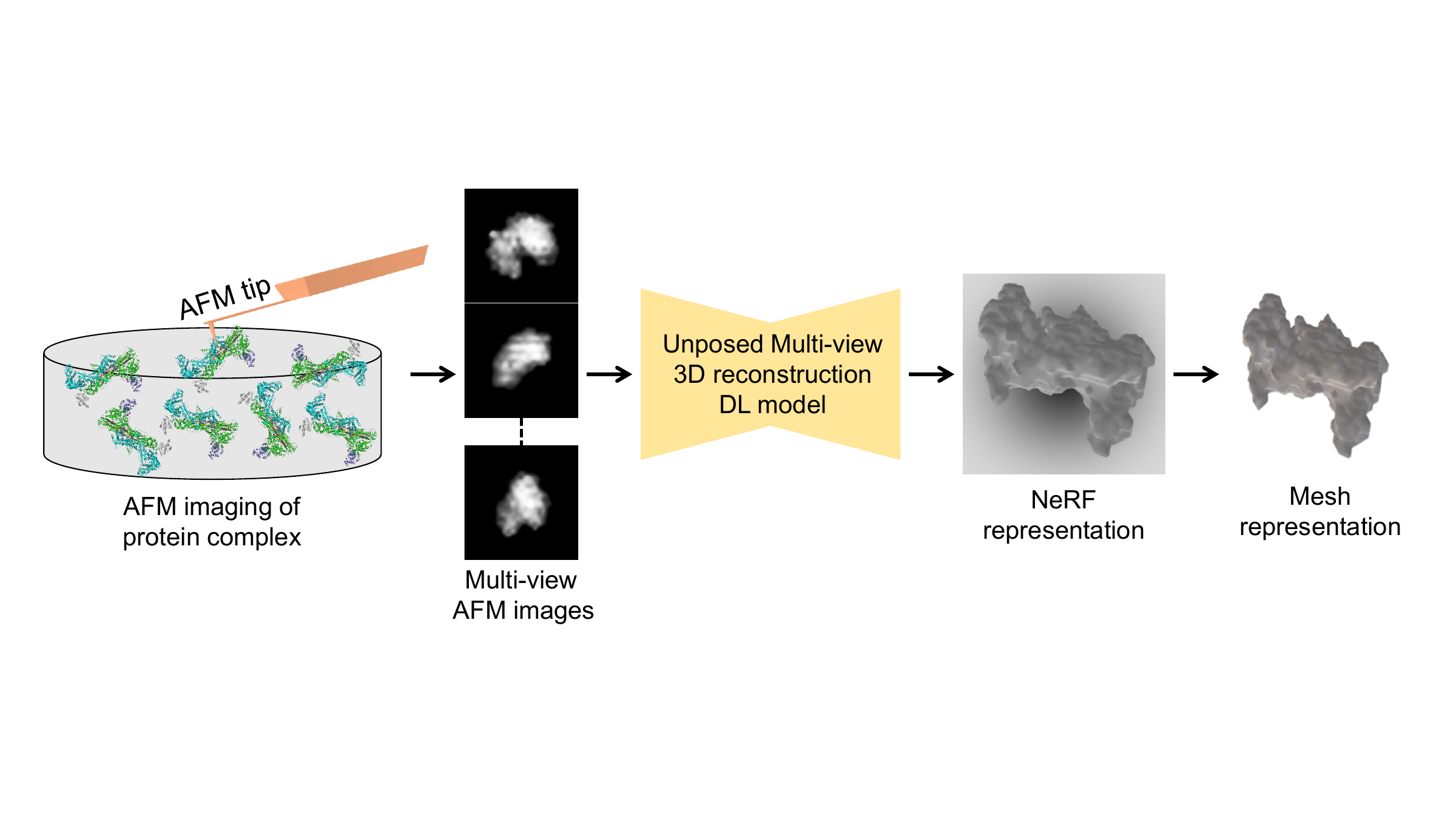}
        \caption{}
        \label{fig:method_overview}
    \end{subfigure}
    \begin{subfigure}[c]{\linewidth}
        \centering
        \includegraphics[width=0.9\linewidth, trim={2in 3.2in 2in 3in},clip]{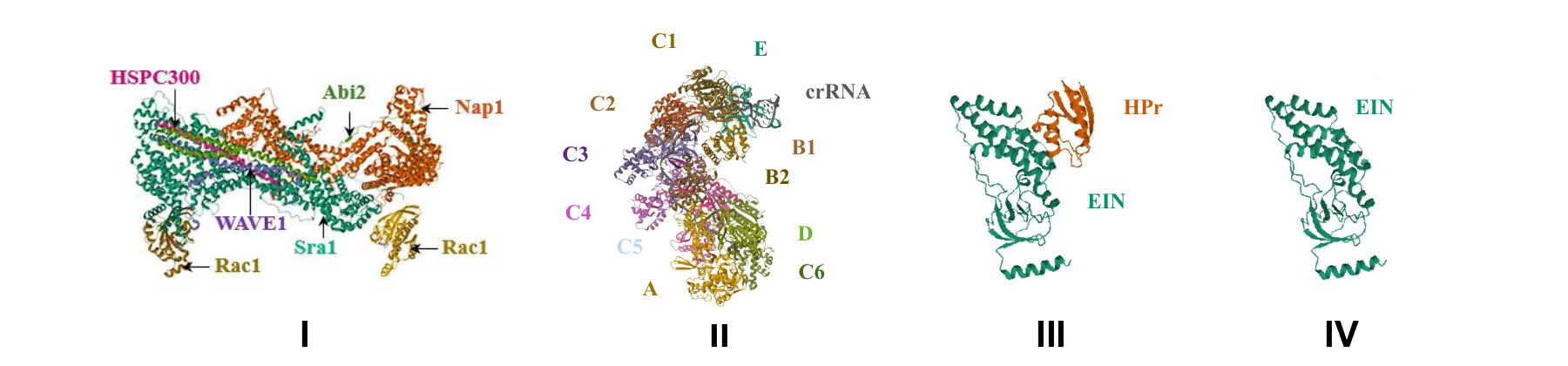}
        \caption{}
        \label{fig:protein_complexes_overview}
    \end{subfigure}
    \caption{(a) Comparative overview of protein structure prediction methods. (Left)
    Experimental techniques like Cryo-EM offer accurate protein structures but are time-consuming and costly. (Right) In contrast, \emph{in silico} approaches such as AlphaFold2-multimer~\citep{AF2Multi2022Evans} and AlphaFold3~\citep{Abramson2024AF3} are often less reliable for large protein complexes and require substantial computational resources for training and inference. (Middle) Our proposed hybrid framework, combining AI with Atomic Force Microscopy (AFM), balances accuracy, experimental cost, and compute resources by leveraging high-resolution AFM images and deep learning to predict 3D protein complex structures more effectively. (b) DL-AFM-based 3D protein reconstruction pipeline. AFM imaging of protein complexes produces multi-view AFM images capturing surface topologies from different views. These images are input to an unposed multi-view 3D reconstruction deep learning model, which outputs a NeRF-based volumetric representation of the protein. The NeRF representation is then converted into a surface mesh for downstream analysis and evaluation. (c) Biological assembly structures of the proteins and protein complexes: (I) Complex structure of WRC-Rac1 (PDB ID: 7USE)~\citep{chen2017rac1}; (II) Type I-E CRISPR-Cascade complex (PDB ID: 5CD4)~\citep{westra2012crispr}; (III, IV) E. coli EIN domain complexed with HPr (PDB ID: 3EZA)~\citep{venditti2015large}.}
    \label{fig:Overview}
\end{figure}

Accurately predicting the 3D structures of proteins and protein complexes (PCs) from their amino acid sequences has remained a significant challenge for many decades~\citep{PF2008Dill}. The protein structure determination is critical, as it directly determines its function~\citep{PF20212Dill}. Over the years, researchers have used and refined experimental techniques such as X-ray crystallography, nuclear magnetic resonance (NMR) spectroscopy, and cryo-electron microscopy (Cryo-EM) to determine the 3D structures of proteins~\citep{NMR2001Kurt, Thompson2020AdvancesIM,cryoEM2015Bai, Jasklski2014ABH}. Cryo-EM single-particle analysis, in particular, has enabled the structural characterization of proteins and large protein complexes that are difficult to crystallize. More recently, integrating deep learning (DL) with cryo-EM has further enhanced structure determination capabilities~\citep{Zhong2021CryoDRGN, He2022ModelAngelo}. However, cryo-EM images are often noisy, with signal-to-noise ratios as low as -20 dB~\citep{Gupta2020MultiCryoGANRO}. Moreover, the experimental process, including grid preparation, screening, and data acquisition, can cost up to \textdollar6000, while downstream computational analysis adds an additional \textdollar1500 per structure~\citep{Cianfroco2015} as of August 2025. Despite these challenges, these experimental advances have greatly expanded the protein structural database.

An alternative to experimental methods involves inferring protein structure directly from amino acid sequences by learning sequence-to-structure relationships. Techniques such as Position Specific Scoring Matrices (PSSM), which quantify mutation frequencies at individual residue positions~\citep{wang2021PSSP}, and Multiple Sequence Alignment (MSA), which captures co-evolutionary relationships across multiple sequences~\citep{Senior2020ImprovedPS, Yang2020}, have played a crucial role in sequence-based structure prediction. Concurrently, advances in DL have transformed the field of protein structure prediction, leading to the development of models such as AlphaFold3 (AF3)~\citep{Abramson2024AF3}, AlphaFold2 (AF2)~\citep{AF22021Jumper}, ESMFold~\citep{esm2fold}, and RoseTTAFold~\citep{RF2021Baek}, many of which rely exclusively on protein sequence information as input. AF2 introduced a novel deep learning architecture, Evoformer, which integrates MSAs and structural templates to predict the 3D structure of individual proteins. Building on the foundation of AF2, AF3 significantly extends these capabilities to model individual proteins and complex biomolecular interactions. Leveraging a diffusion-based generative architecture~\citep{karras2022edm}, AF3 can jointly model structures involving proteins, nucleic acids (DNA and RNA), small molecules, ions, and modified residues. Despite these advancements, the accuracy of such models decreases when predicting large protein complexes where multiple proteins interact in close proximity. For example, AlphaFold2-Multimer~\citep{AF2Multi2022Evans}, designed explicitly for protein complexes prediction, and AF3 fails to reliably predict the structure of the WAVE Regulatory Complex (WRC-Rac1)~\citep{koronakis2011wave, rottner2021wave, ismail2009wave, chen2014wave}, as illustrated in \figref{fig:motivation_overview}. This limitation likely arises from the inherent difficulty of predicting large protein complex structures based solely on amino acid sequences without incorporating complementary experimental data.

Given the limitations of current \emph{in silico} methods, which often struggle to accurately predict PC structures, and the high cost and time requirements of experimental techniques such as Cryo-EM, we propose a novel hybrid approach. Our method leverages DL models trained on images acquired via Atomic Force Microscopy (AFM), as illustrated in \figref{fig:method_overview}. We formulate the protein structure prediction as a multi-view 3D reconstruction task using AFM images. AFM~\citep{sarkar2019live,sarkar2015interaction,sarkar2022biosensing,sarkar2022multimodal,stanley2023effects,jones2017revisiting}, a type of scanning probe microscopy, provides a non-destructive, high-resolution imaging modality that can visualize proteins under near-physiological conditions. Unlike other experimental techniques requiring extensive sample preparation, such as freezing, drying, or fluorescent labeling, AFM preserves the integrity and native state of protein structures. Importantly, the cost of AFM-based data collection for a protein sample, including sample preparation and incubation, is approximately \textdollar500 as of August 2025, which is substantially lower than the expenses associated with Cryo-EM. Predicting the structures of large PCs remains challenging for sequence-based methods because they lack explicit 3D spatial cues of interacting subunits. AFM uniquely addresses this gap. When proteins adsorb onto the substrate, they adopt random orientations, naturally producing multiple 2D projections of the PCs. These multi-view images provide complementary spatial information about the overall topology of the PC that purely sequence-based methods cannot capture. By leveraging this multi-view property, AFM offers a promising avenue for reconstructing 3D structures of challenging PCs. 

However, training a DL model for this 3D reconstruction task requires a large dataset of proteins with corresponding AFM images, which is impractical to obtain experimentally, and no such large-scale public datasets currently exist. To overcome this limitation, we introduce a virtual AFM framework, a rasterization-based simulation pipeline that mimics the AFM imaging process and synthetically generates multi-view AFM images. This approach enables the creation of large, diverse datasets by generating multi-view AFM images for a wide range of proteins and PCs, effectively addressing the data scarcity challenge.

While our virtual AFM framework overcomes the data limitation, reconstructing 3D structures from AFM images introduces an additional challenge. The experimental AFM imaging process does not provide the pose or orientation of each protein molecule. Most 3D reconstruction methods rely on known camera poses or alignment cues or approximate poses for input views to recover geometry from multi-view inputs~\citep{Mildenhall2021NeRF, Yu2020FastMVSNetSM, lin2021barf, goel2022differentiable, chen2024multiview}, which is not feasible in AFM imaging. We build ProFusion, a protein structure prediction model based on the UpFusion architecture~\citep{kani2024upfusion} to address this. UpFusion is specifically designed to synthesize novel views and reconstruct 3D shapes from unposed images, demonstrating strong generalization beyond training categories. Using our large-scale virtual AFM dataset, ProFusion is trained to directly predict the 3D structures of proteins and PCs from multi-view AFM images without requiring input pose supervision. We explain the details of the architecture and training process in \secref{sec:profusion}. We extensively validate our method on virtual and experimental AFM images of multiple PCs, including CRISPR-Cascade~\citep{westra2012crispr, kiro2014efficient}, EIN-Hpr~\citep{venditti2015large} from Enzyme I, and WRC-Rac1~\citep{chen2017rac1}. These complexes span a diverse range of structural architectures. The CRISPR-Cascade (CasB-E) complex forms a curved, elongated assembly composed of multiple subunits arranged along a crRNA backbone (seahorse-shaped) ~\citep{westra2012crispr, jackson2014crystal}. The WRC-Rac1 complex adopts an extended, multi-lobed conformation due to its flexible structural arrangement ~\citep{chen2017rac1, stovold2005inclusion, rottner2021wave}. Conversely, the EIN-HPr complex displays a compact structure formed by two interacting proteins aligned side by side ~\citep{venditti2015large, doucette2011alpha}. We explain the details of each protein complex in \secref{sec:protein_complexes}.

In summary, this work presents a hybrid framework that combines an \emph{in silico} DL model with experimental AFM imaging for 3D protein structure prediction. We develop a virtual AFM data generation pipeline to create a multi-view AFM dataset for proteins, which we use to train a deep learning model to reconstruct 3D protein structures from unposed images. We validate the approach on both virtual and experimental AFM images. Our key contributions are as follows:
\begin{itemize}
    \item We propose a novel, cost-effective method for predicting protein and protein complex structures using deep learning models, trained on AFM images.
    \item We introduce a \textit{virtual} AFM framework and generate a large-scale synthetic dataset comprising approximately 542,000 proteins with corresponding multi-view AFM images.
    \item  We validate our approach using experimental AFM images across multiple protein complexes, demonstrating its practical effectiveness.
\end{itemize}

\section{Methods}\label{methods}

\subsection{AFM Setup}

Atomic Force Microscopy (AFM) is a powerful imaging technique that provides high-resolution and 2.5D surface profiles of samples at the nanometer scale. It operates by scanning a sharp, flexible probe across the sample surface, which is mounted on a micron-scale cantilever. AFM measures the interaction forces between the probe and the sample surface to characterize the sample. As the probe scans across the sample, variations in these forces cause deflections in the cantilever, which are detected by a laser beam reflected from the cantilever into a photodetector. This deflection data is used to construct detailed topographical maps of the sample (see Supplementary Material, \figref{fig:AFM setup}) ~\citep{cai2018atomic, deng2018application, mcclelland1987atomic}. 

\subsubsection{AFM Sample Preparation}

AFM-compatible mica samples are prepared by adhering muscovite mica discs (50-15, Highest Grade V1 AFM Mica Discs, 15 mm, Ted Pella) to stainless steel specimen support discs (20 mm RS-MN-40-100020-50, Rave Scientific). These samples are left on the benchtop inside a 100 X 15 mm plastic dish to air dry for 24 hours. Next, a $300\mu$l solution of the protein or protein complex is prepared using the appropriate buffer at various concentrations. The muscovite mica disc is freshly cleaved to ensure an extremely flat and negatively charged surface, making it suitable to adhere the coated protein solution for AFM imaging. Immediately following cleaving, the protein solution is carefully applied to the mica substrate and incubated at 4°C for 24 hours. After incubation, the substrate is rinsed with the appropriate buffer and incubated in 200 $\mu$l fresh buffer solution before securing the sample on the AFM stage using magnetic contact. The AFM experiment is conducted in fluid mode. Detailed workflow for the AFM sample preparation is shown in \figref{fig:Sample preparation}.

This study utilizes the BioScope Resolve system (Bruker) for AFM imaging. A ScanAsyst-Fluid probe and a PEAKFORCE-HIRS-F-A probe with silicon nitride tips and reflective gold coatings on the back, featuring spring constants of 0.70 N/m and 0.4 N/m, respectively, and tip radius of 20 nm and 1.5 nm, respectively, are employed for scanning and visualizing the 2.5D surface profiles of the protein and protein-complex samples. AFM imaging is conducted in PeakForce QNM mode in fluid, within a vibration- and noise-isolated chamber to minimize random noise and environmental interference in the data. The deflection sensitivity and cantilever spring constant for each probe used in the AFM experiment are calibrated before each experiment. Additionally, high-resolution imaging is performed with a resolution of 256 × 256 pixels per line, and the peak force frequency is maintained at 1 kHz throughout the characterization process.

\subsubsection{Protein Complexes}\label{sec:protein_complexes}

To validate our work with AFM-based 2.5D protein and protein-complex images, we have used user cases: CRISPR-Cascade~\citep{westra2012crispr}, EIN-Hpr~\citep{venditti2015large} from Enzyme I, and WRC-Rac1~\citep{chen2017rac1}. Clustered regularly interspaced short palindromic repeats (CRISPR) and the CRISPR-associated complex (Cascade) function as an antiviral defense system. The CRISPR-Cas system is like an immune system for bacteria and archaea, protecting them from invaders like viruses and foreign genes by using a guide CRISPR RNA (crRNA) molecule to recognize them~\citep{westra2012crispr}. 
The Cascade structure consists of 11 protein subunits and a single crRNA, making a total of 12 subunits in the complex~\figref{fig:protein_complexes_overview}(II) \citep{zhao2014crystal}. The Cascade structure comprises the following subunits: CasA, CasB (B1-B2), CasC (C1-C6), CasD, CasE, and crRNA. Cascade is a dsDNA (double-stranded DNA) binding protein that uses a guide crRNA to bind to complementary DNA~\citep{zhao2014crystal, jackson2014crystal}. Following the protocols used for Cas protein use case, we immobilize and visualize Cas B-E using bio-AFM (CasB, CasC, CasD, CasE, and crRNA), which are amplified from E. coli K12 (details purification process are mentioned in the supplementary \secref{sec:protein_purification}).

Enzyme I is a key component of the bacterial phosphotransferase system (PTS), responsible for the initial phosphorylation of sugar molecules during their transport across the cell membrane. EIN is the N-terminal domain of Enzyme I, a crucial component of the bacterial phosphotransferase system (PTS) involved in sugar transport and metabolism in bacteria. Along with EIN, HPr (histidine-containing phosphocarrier protein) plays a key role in this system ~\citep{venditti2015large, doucette2011alpha}. Enzyme I (EI) is composed of two domains: two N-terminal phosphoryl transfer domains (EIN) on both sides (that house the PEP binding site) and one C-terminal dimerization domain (EIC) in the middle. The isolated EIN domain can reversibly transfer a phosphoryl group to HPr. The EIN domain is further divided into two subdomains that contain the active site histidine (His189), and its interaction surface for HPr is shown in the schematic figure of \figref{fig:protein_complexes_overview}(III, IV). Both EIN and HPr are derived from Escherichia coli ~\citep{schwieters2010solution, teplyakov2006structure, takayama2011combined}. In this work, EIN and HPr are expressed and purified as described by ~\citet{nguyen2018oligomerization}. We individually image EIN using bio-AFM first and combine the EIN and Hpr to make the EIN-Hpr complex combination, followed by our imaging of 2.5D structures of that protein complex.

The WASP (Wiskott-Aldrich Syndrome Protein) family WAVE (WASP-family verprolin-homologous protein) regulatory complex (WRC) and Rac1 (member of the Rho family of small GTPases) are integral components in the regulation of the actin cytoskeleton that is a dynamic network crucial for cell shape, movement, and division ~\citep{chen2017rac1, kurisu2009wasp}. Each WAVE protein is integrated into a heteropentameric structure called the WAVE Regulatory Complex (WRC) in cells. WRC itself is a complex protein structure of the Sra1, Nap1, Abi2, and HSPC300, or their respective homologs (\figref{fig:protein_complexes_overview}(I)). The WRC complex functions as a guanine nucleotide exchange factor (GEF) for Rac1, facilitating its activation. This interaction is pivotal for various cellular processes, including cell migration, morphology, and proliferation ~\citep{chen2017rac1, stovold2005inclusion, rottner2021wave}. In this work, the protein complexes are expressed and purified as described by ~\citet{ding2022structures}, and we visualize the WRC-Rac1 complex using bio-AFM to observe the protein-complex structure.

\subsubsection{AFM Imaging and Scan Parameters}\label{sec:afm_scan_params}

Different optimized scanning parameters are used for different proteins and protein complexes to prevent damage to the soft protein samples and the AFM probes, and to ensure clear and detailed imaging of topographical structures. We explore different scanning areas ranging from 350 nm to 1000 nm with an image resolution of $256 \times 256$, yielding per pixel resolution from 1.37 nm/pixel to 3.90 nm/pixel. The parameters used for different samples are as follows: 

\textbf{Cas B-E:} We use deflection sensitivity of 20.98 nm/V, cantilever spring constant of 0.28 N/m, scan rate of 0.343 Hz, peak force setpoint at 720 pN, and peak force amplitude of 300 nm. The appropriate buffer used is a size exclusion buffer (20 mM Tris-HCl, 100 mM KCl, and 5\% glycerol in 1X concentration). The sample solution is successfully concentrated to 0.5 $\mu$M and 10 $\mu$M.

\textbf{EIN-HPr:} The deflection sensitivity for EIN is 22.94 nm/V, with a corresponding cantilever spring constant of 0.37 N/m, a scan rate of 0.374 Hz, a peak force setpoint of 650 pN, and a peak force amplitude of 300 nm. For the EIN-HPr complex, the deflection sensitivity is 22.66 nm/V, with a spring constant of 0.27 N/m, a scan rate of 0.415 Hz, a peak force setpoint of 570 pN, and the same peak force amplitude of 300 nm. For experimenting, EIN is imaged in a buffer containing 50 mM Tris-HCl and 150 mM NaCl at pH 8, with sample concentrations of 3.7 $\mu$M and 10 $\mu$M tested across trials. The EIN-HPr complex is also prepared in the same Tris buffer (50 mM Tris-HCl, 150 mM NaCl, pH 8), but at a higher concentration of 25 $\mu$M to ensure stable complex formation and effective immobilization on the substrate.

\textbf{WRC-Rac1:} The calibrated cantilever spring constant is 0.70 N/m. We use a scan rate of 1.00 Hz and a peak force amplitude of 250 nm. The appropriate buffer and successful sample concentration for WRC-RAC1 are prepared as follows: Buffer B, consisting of 10 mM HEPES, pH 7, 2 mM  MgCl$_2$, 100 mM NaCl, 10\% glycerol, and 1 mM DTT, with a sample concentration of 0.005 $\mu$M.

\subsection{ProFusion: Deep Learning Model Architecture}\label{sec:profusion}
\begin{figure*}[t!]
    \centering
    \begin{subfigure}[t]{\linewidth}
    \centering
    \includegraphics[width=0.86\linewidth, trim={1.9in 2.2in 2.15in 1.5in},clip]{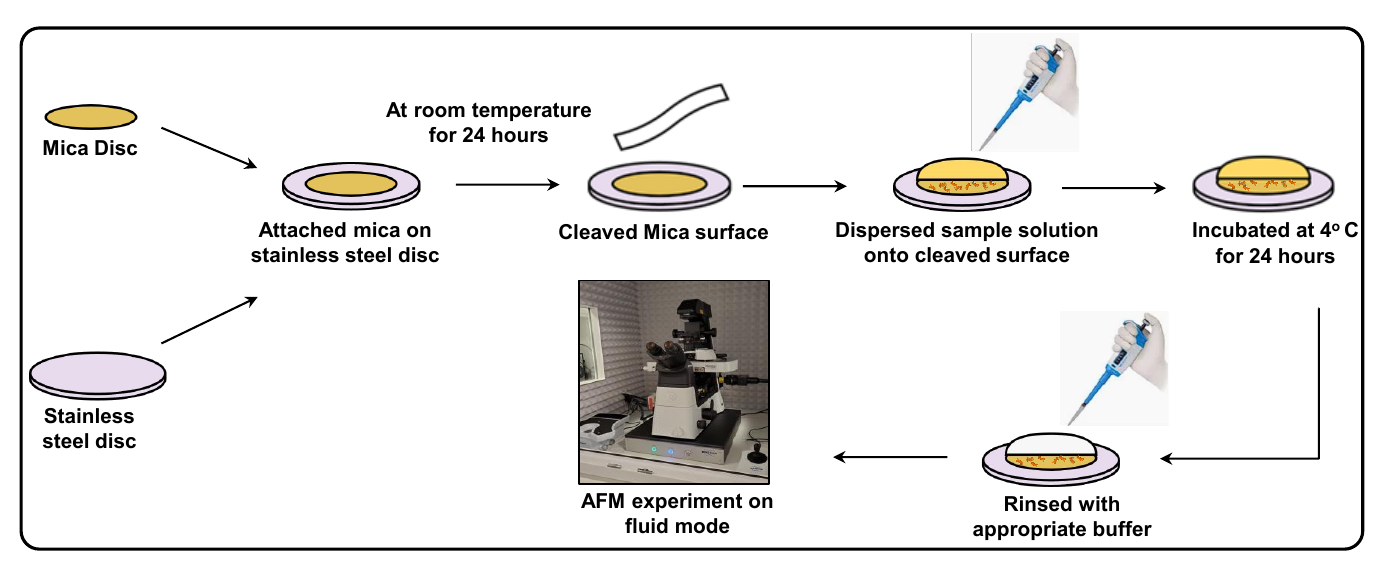}
    \caption{}
    \label{fig:Sample preparation}
    \end{subfigure}
    \begin{subfigure}[t]{\linewidth}
    \centering
    \includegraphics[width=0.86\linewidth, trim={0.25in 1.15in 0.25in 1.15in},clip]{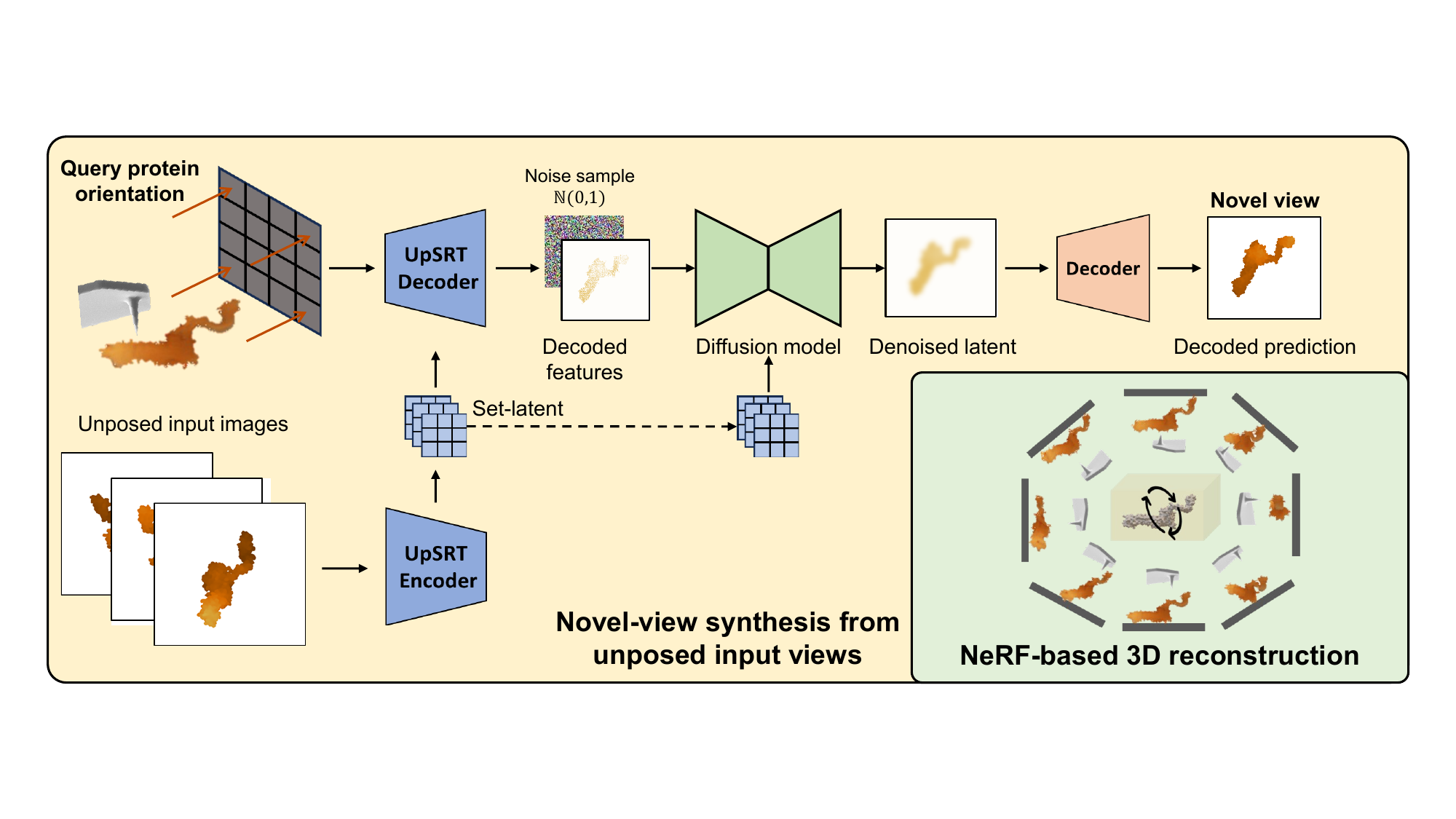}
    \caption{}
    \label{fig:upfusion_architecture}
    \end{subfigure}
    \begin{subfigure}[t]{\linewidth}
    \centering
    \includegraphics[width=0.86\linewidth, trim={0in 1.5in 0in 1.5in},clip]{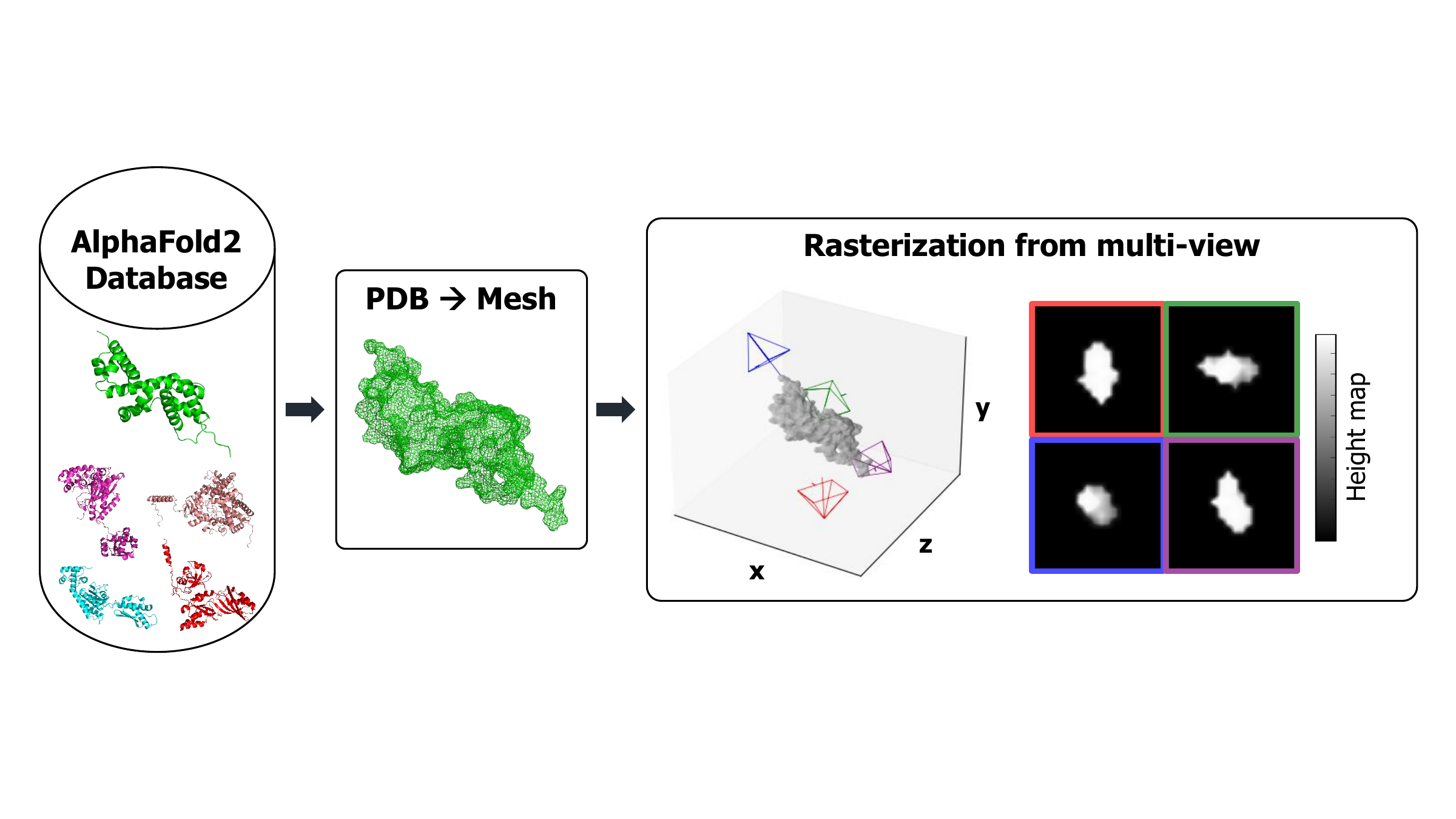}
    \caption{}
    \label{fig:virtual_afm}
    \end{subfigure}
    \caption{(a) Workflow for preparing AFM-compatible samples. Muscovite mica discs (15 mm, highest-grade V1) are mounted on stainless steel supports and dried at room temperature for 24 h. The mica is freshly cleaved to expose the flat and negatively charged surface, and $300\mu$l of protein or protein complex solution in the appropriate buffer is immediately deposited. Samples are incubated at 4 °C for 24 h, rinsed with fresh buffer, and immersed in 200 $\mu$l buffer before being mounted on the AFM stage for imaging in fluid mode. (b) Overview of our UpFusion-based pipeline for novel view synthesis and 3D NeRF reconstruction of protein complexes. Step I: Novel View Synthesis from unposed multi-view AFM images using the UpSRT encoder-decoder and a conditional diffusion model. Step II: Train an instance-specific NeRF for 3D reconstruction of protein structure. (c) An overview of the Virtual AFM framework used to synthetically generate multi-view AFM-like images from 3D protein structures. Protein structures are obtained from the AlphaFold2 Database in PDB format and converted into 3D surface meshes. These meshes are rendered from multiple camera viewpoints using a rasterization-based process simulating AFM imaging. The resulting depth maps are inverted to produce height maps. }
\end{figure*}

Reconstructing 3D structures from sparse, unposed 2D images is particularly challenging in scenarios like AFM experiments, where camera poses are unknown or difficult to estimate. Most existing sparse-view 3D reconstruction methods rely on geometric aggregation of input views with known camera poses~\citep{Yao2018MVSNet, yu2021pixelnerf, Chan2023GenerativeNV, zhou2023sparsefusion}, which limits their applicability to experimental acquired AFM images. To enable direct 3D protein structure prediction from unposed multi-view AFM images, we introduce ProFusion, built by adapting the UpFusion architecture~\citep{kani2024upfusion}.
As shown in \figref{fig:upfusion_architecture}, UpFusion performs two key tasks: novel view synthesis from sparse, unposed images and inference of 3D object representations from synthesized views. It achieves this by leveraging a conditional generative model like diffusion models~\citep{Ho2020ddpm,nichol2021iddpm,Rombach2022ldm} to synthesize novel views for different arbitrary query poses, guided by the input images as context. This eliminates the need for explicit input camera pose supervision. Finally, it optimizes an instance-specific neural radiance field (NeRF)~\citep{Mildenhall2021NeRF} network using the generated novel views for the given query poses to produce the final 3D representation.

For the novel view synthesis task, it integrates two key components: a scene-level transformer and a conditional diffusion model. The transformer, implemented as the Unposed Scene Representation Transformer (UpSRT)~\citep{Mehdi2022UpSRT}, infers features aligned to the query view by implicitly leveraging all available input images as contextual information. While UpSRT alone can generate novel views, its outputs tend to be blurry due to its training objective, which relies solely on mean-squared error for pixel-wise RGB reconstruction. To address this limitation, UpFusion introduces a latent diffusion model~\citep{Rombach2022ldm} conditioned on the decoder features produced by UpSRT, enabling probabilistic and sharper novel view synthesis~\citep{Rombach2022ldm, zhang2023controlnet}. To further enhance visual fidelity, it incorporates shortcut connections via attention mechanisms, allowing the diffusion model to directly access encoder-level features from UpSRT during generation, thereby preserving finer details in the synthesized views. Rather than training a large latent diffusion model from scratch, we use the ControlNet architecture~\citep{zhang2023controlnet} to adapt a pre-trained Stable Diffusion v1.5 model~\citep{Rombach2022ldm} for handling AFM images.

While the conditional diffusion model is effective at generating high-quality novel views, it does not inherently guarantee 3D consistency across these views. To address this, we train an instance-specific NeRF network for predicting 3D structure. Specifically, it optimizes an Instant-NGP framework~\citep{mueller2022instant, torch-ngp}, which enables fast and memory-efficient training. This optimization is guided by Score Distillation Sampling (SDS)~\citep{poole2023dreamfusion}, a technique that leverages the pretrained diffusion model to supervise the training of the NeRF network. The core idea behind SDS is to distill the generative knowledge of the conditioned diffusion model to obtain a 3D representation by guiding the NeRF network to produce renderings that match with those synthesized by the diffusion model for the same query poses~\citep{zhou2023sparsefusion}. This step ensures that the final 3D structure is visually plausible from individual viewpoints and geometrically consistent across all views. 

\subsubsection{Training Details}

To develop ProFusion, we train the UpFusion architecture on a large-scale dataset of synthetically generated virtual AFM images of proteins and PCs using a three-stage procedure. The input to ProFusion consists of grayscale AFM images with a resolution of $256 \times 256$ pixels.

Stage 1 – UpSRT training. We train the UpSRT from scratch to predict a set of query views given a set of ground-truth reference images, using a mean squared error (MSE) reconstruction loss. Training is performed on 8 NVIDIA A100 GPUs (40 GB VRAM each) across 2 nodes, with a global batch size of 32 (32 protein samples, each with 6 input AFM images). We use the Adam optimizer~\citep{kingma2014adam} with a learning rate of $6\times10^{-5}$ for $\sim$32 K optimization steps, requiring ~176 GPU-hours.

Stage 2 – Diffusion model fine-tuning. After training UpSRT, we freeze its weights and pass its decoder features as conditioning inputs to a ControlNet architecture, fine-tuning a pre-trained Stable Diffusion v1.5 model~\citep{Rombach2022ldm} for AFM image synthesis. This stage uses a global batch size of 256 on the same 8 × A100 GPU configuration across 2 nodes, trained with Adam optimizer with learning rate $6\times10^{-5}$ for $\sim$25 K steps, totaling ~384 GPU-hours.

Stage 3 – 3D representation optimization. Using the trained diffusion model, we obtain an instance-specific 3D protein structure by optimizing an Instant-NGP~\citep{mueller2022instant, torch-ngp} NeRF with the SDS approach discussed above. We train for 3000 iterations (~45 minutes) for each protein sample on a single NVIDIA A100 GPU, producing a geometrically consistent 3D structure that matches the diffusion model’s synthesized views. Once the NeRF is optimized, we extract the 3D surface mesh by first converting the learned density field from NeRF representation into an occupancy grid, then applying the Marching Cubes algorithm~\citep{lorensen1987marchingcubes} to generate a mesh, which is then used for quantitative evaluation against ground-truth protein structures.

\subsection{\textit{Virtual} AFM}
\label{sec:virtualAFM}

Training a neural network typically requires a large dataset of paired input-output examples. Protein structure reconstruction from AFM images needs a dataset comprising multiple AFM images of each protein, capturing different views paired with corresponding 3D structural representations. However, such a dataset is not publicly available, and collecting one experimentally via AFM imaging is extremely time-consuming and resource-intensive, likely taking years to complete.

We developed the \textit{Virtual} AFM framework to address this limitation. \textit{Virtual} AFM uses rasterization, a computer-graphics technique, to simulate the AFM imaging process to generate AFM-like images from known 3D structures of proteins. Rasterization method expects an input in mesh format. To facilitate this, we use PyMol, a widely adopted tool for protein visualization, to export protein structures from PDB format into 3D mesh files in OBJ format. PyMol supports various rendering styles, including surfaces, spheres, sticks, cartoons, and ribbons. For our framework, we export the surface mesh representation in OBJ format, as it best approximates the physical topology captured by AFM.

\begin{algorithm}[t!]
    \caption{\textit{Virtual} AFM Pipeline}\label{Alg:virtual_afm}
    \SetKwInOut{Input}{Input}
    \SetKwInOut{Output}{Output}

    \Input{PDB file of a protein structure, scanning step size, AFM tip radius, $N$: number of desired views}
    \Output{$N$ virtual AFM images and corresponding height maps (in nm units)}
    \setlist[enumerate]{left=0pt} 
    \begin{enumerate}
        \item Convert the PDB file into a 3D surface mesh (in nm units) using PyMol.
        \item Normalize and center the mesh within a unit cube with bounds $[-1, -1, -1]$ to $[1, 1, 1]$; store the applied scale factor.
        \item Compute the image resolution based on the given scanning step size.
        \item Initialize the PyTorch3D rasterizer with appropriate camera and rasterization settings. 
        \end{enumerate}
        \For{$i=0:N$}
        {
        \setlist[enumerate]{left=0pt}
        \begin{enumerate}
        \item Randomly sample a camera pose: 
        elevation $\in [-90^\circ, 90^\circ]$, azimuth $\in [0^\circ, 360^\circ]$, and roll $\in [0^\circ, 360^\circ]$.
        \item Render the depth map from the sampled pose using rasterization.
        \item Invert the depth values to obtain the height map.
        \item Apply morphological dilation using a kernel determined by the AFM tip radius to simulate tip convolution.
        \item Rescale the height map using the stored scale factor to express values in nanometers.
        \item Save the height map and export the corresponding rendered view as a PNG image.
        \end{enumerate}
        }
\end{algorithm}

AFM generates topographic images by scanning the surface of a protein sample using the AFM tip, producing detailed height maps that reflect surface topology. In our \textit{Virtual} AFM framework, we simulate this scanning process by employing a rasterization technique that converts 3D geometry into 2D images by projecting the visible surfaces onto an image plane. Conceptually, we treat the virtual camera as a proxy for the AFM tip. It observes the protein surface from its orientation and records depth information along its viewing direction, much like how the AFM tip measures height by probing the surface point-by-point. We implemented this using PyTorch3D~\cite{ravi2020pytorch3d} to render depth maps from the 3D surface mesh of the protein. Rasterization assigns each pixel in the image a depth value representing the distance from the camera (i.e., the virtual AFM tip) to the closest visible surface. To match the convention of AFM images, where brighter pixels represent higher surface points, we invert the depth maps to produce height maps analogous to those captured by experimental AFM scans.

To mimic the random orientations of protein molecules adsorbed onto an AFM substrate, we randomly sample virtual camera positions on a unit sphere by varying elevation, azimuthal, and roll angles. This allows us to simulate AFM imaging from multiple viewpoints, generating diverse, multi-view virtual AFM images for each protein structure as demonstrated in \figref{fig:virtual_afm}. Our method also incorporates key experimental factors, including contact-based height detection, scanning resolution, and tip convolution arising from the AFM probe's shape and size. Contact-based height detection is simulated via rasterization as discussed above. We model scanning resolution by adjusting the image resolution used during rasterization. We apply morphological dilation to the rendered height maps to model tip convolution effects. The complete pipeline for the \textit{Virtual} AFM framework is summarized in \coloredref{Algorithm}{Alg:virtual_afm}.

We use the AlphaFold DB~\citep{Varadi2021AFDB}, a repository of structural predictions for Swiss-Prot entries generated by the AlphaFold2 model~\citep{AF22021Jumper}. Although AlphaFold2 faces challenges in accurately predicting protein complexes, it performs exceptionally well for individual protein structures. We used this resource to generate a large-scale virtual AFM dataset comprising approximately 542,000 proteins and their corresponding multi-view virtual AFM images. To ensure structural reliability, we filtered the dataset by applying a threshold on the average predicted Local Distance Difference Test (pLDDT) score for each structure. Only proteins with an average pLDDT score of 80 or higher were retained, resulting in a high-confidence subset of around 448,000 protein structures that are used for training DL models. This dataset provides a rich resource for training DL models across various downstream applications in protein structure analysis and AFM-based imaging.

\section{Results}\label{results}

\subsection{Protein Structure Imaging Using AFM}

\begin{figure}[h!]
    \centering
    \begin{subfigure}[c]{\linewidth}
        \centering
        \includegraphics[width=0.74\linewidth, trim={0.0in 5.1in 0.0in 0.0in},clip]{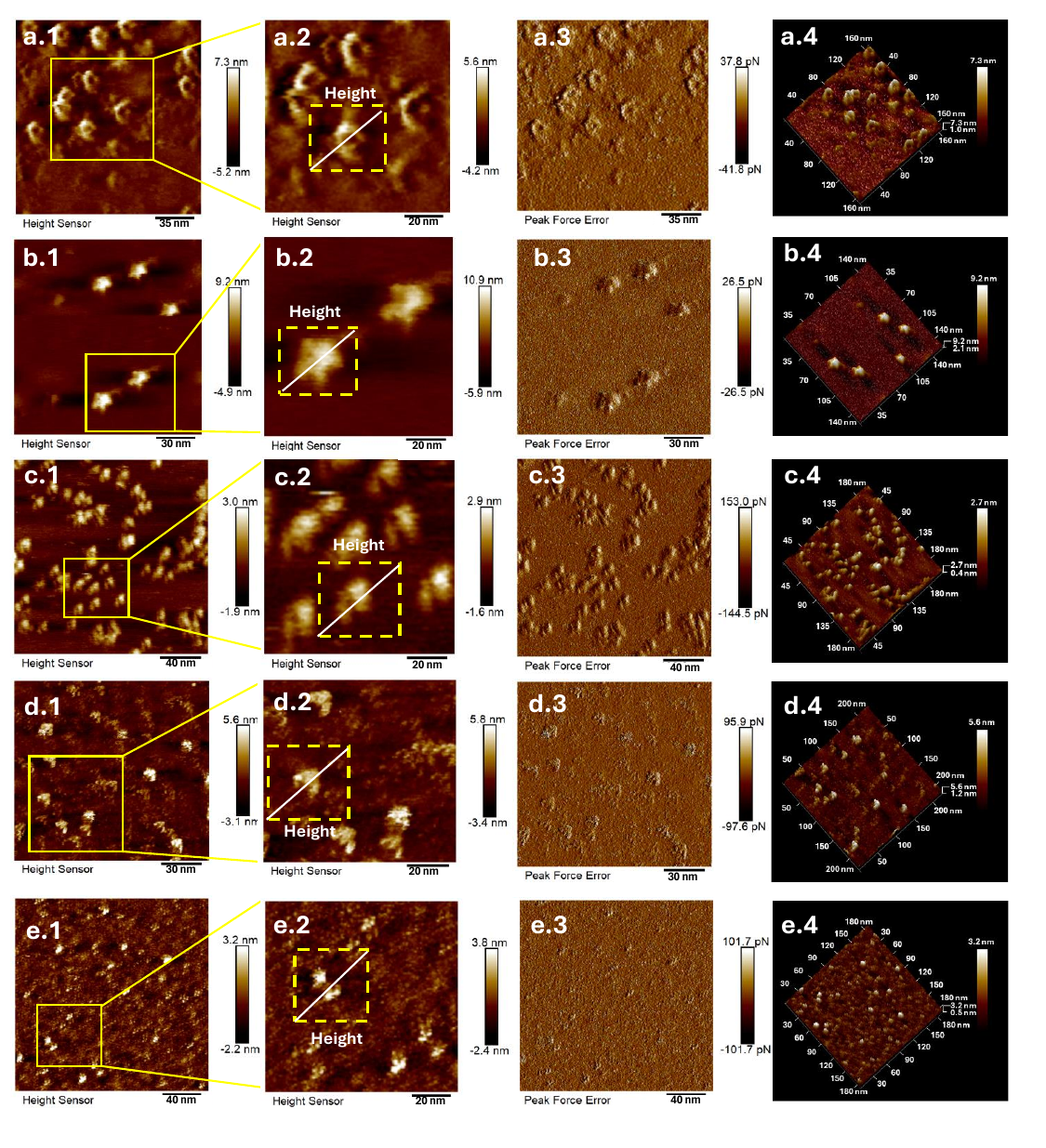}
        \label{fig:AFM height and zoomed images of proteins}
    \end{subfigure}
    \begin{subfigure}[c]{\linewidth}
        \centering
        \includegraphics[width=0.5\linewidth]{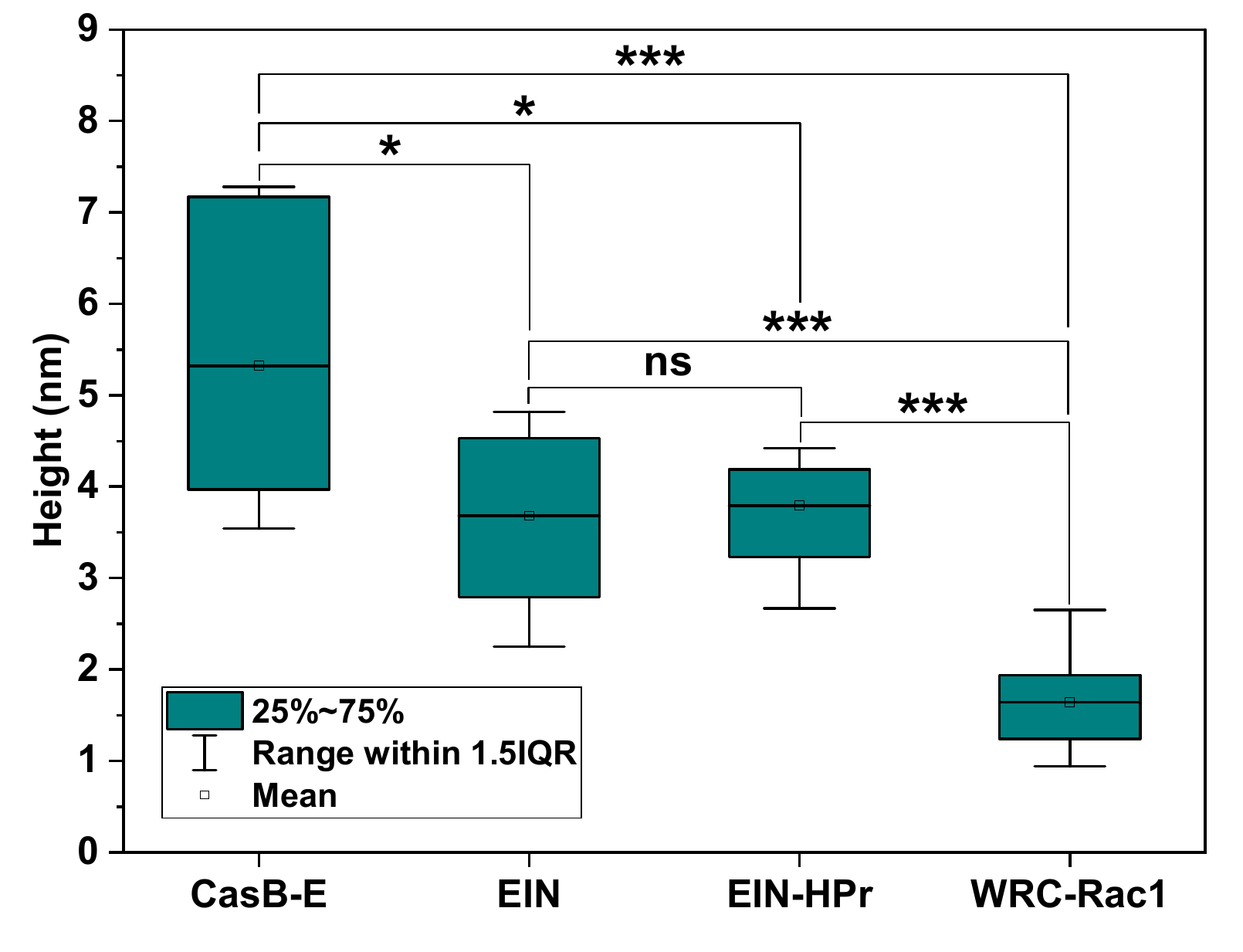}
        \label{fig:Height R1}
    \end{subfigure}
    \caption{High-resolution AFM images and quantitative analysis of protein and protein complex structures. (a–e) show four imaging modes for each sample: (1) height sensor image offering an overview of the surface, (2) zoomed-in height sensor image with a marked line for height estimation, (3) peak force error map highlighting mechanical contrast, and (4) 2.5D topographic view. Samples include: (a) CasB-E complex with a seahorse-like shape, (b) CasB-E complex with a star-like shape, (c) WRC-Rac1 complex, (d) EIN-HPr protein complex, and (e) EIN protein. (f) Quantitative analysis of height measurement. CasB-E shows the highest average height (5.32 nm), while WRC-Rac1 has the lowest (1.64 nm). All measurements are obtained using NanoScope Analysis software 3.00. Statistical significance is denoted as follows: *~\textit{p}~$<$~0.05,\quad **~\textit{p}~$\leq$~0.01,\quad ***~\textit{p}~$<$~0.0001,\quad \textit{p}~$\geq$~0.05 = not significant (ns).}
    \label{fig:AFM height and zoomed images of proteins}
\end{figure}

High-resolution AFM imaging reveals multi-view topographical structures for each protein and protein complex (\figref{fig:AFM height and zoomed images of proteins}), where the height sensor image offers an overview of the surface, the peak force error map highlights mechanical contrast and provides a better view, and the 2.5D topographic view offers height and contrast together. In an earlier study, the CRISPR-Cascade sample (CasB-E complex) was readily identified by its seahorse-shaped profile in the topographic views~\citep{jackson2014crystal}. However, due to randomly oriented CasB-E attachment on the mica surface, we observe a consistent curved head-and-tail architecture characteristic of the Cascade complex (\figref{fig:AFM height and zoomed images of proteins}{a}). In some orientations, the complex appears slightly more compact or multi-lobed (star-like in outline, \figref{fig:AFM height and zoomed images of proteins}{b}). However, the overall curved contour remains evident, matching the known Cascade seahorse-shaped structure (\figref{fig:protein_complexes_overview} II). Also, quantifying its height profile presents the highest vertical dimension among the samples (height: 5.32 nm average), reflecting its large multi-subunit assembly (\figref{fig:AFM height and zoomed images of proteins}{f}).
AFM images of WRC-Rac1 (\figref{fig:AFM height and zoomed images of proteins}{c}) show a broad shape with multiple protrusions, suggesting the presence of its subunits. Despite some variation with orientation, the multi-view images consistently demonstrate an extended topology rather than a compact globule. This corresponds well with the extended conformation of the WAVE Regulatory Complex bound to Rac1 (\figref{fig:protein_complexes_overview} I). Notably, WRC-Rac1 has the smallest apparent height (1.64 nm average), indicating it lies relatively flat on the substrate. However, its lateral shape is still the key indicator of the structure obtained via AFM imaging.

For the EIN protein (the N-terminal domain of Enzyme I), AFM imaging shows a compact topographic shape (\figref{fig:AFM height and zoomed images of proteins}{e}). EIN molecules adsorbed on the surface appear roughly longish, consistent with a single-domain protein. In contrast, when EIN was complexed with HPr protein, the resulting EIN-HPr complex displays a clearly elongated shape in the AFM images (\figref{fig:AFM height and zoomed images of proteins}{d}). This elongation reflects the HPr protein bound to EIN, forming a joint structure in which the two components lie adjacent to each other. However, the average height of the EIN-HPr complex (3.8 nm) is slightly higher than EIN alone (3.7 nm), indicating that HPr binds alongside EIN rather than stacking vertically, thereby possibly increasing the lateral size more than the height (\figref{fig:AFM height and zoomed images of proteins}{f}). Additionally, surface roughness measurements of these samples can reveal more information about their surface irregularities at the nanoscale, where we find the roughness measurements ranged from 0.24 nm for the smoothest sample (WRC-Rac1) up to 1.19 nm for the roughest (EIN). Please refer to Supplementary Material,  \figref{fig:Roughness R1}. Furthermore, we performed statistical comparisons among the samples using t-tests to evaluate differences in height, as illustrated in \figref{fig:AFM height and zoomed images of proteins}f. For height, significant differences were found between CasB-E and EIN (\textit{p}~$<$~0.05), CasB-E and EIN-HPr (\textit{p}~$<$~0.05), and even higher differences between CasB-E and WRC-Rac1 (\textit{p}~$<$~0.0001). No significant differences were observed between EIN and EIN-HPr, indicating similar vertical profiles. However, both EIN and EIN-HPr were significantly higher than WRC-Rac1 (\textit{p}~$<$~0.0001).

\subsection{Protein Structure Prediction Performance of DL Model}

\begin{table*}[ht!]
\centering
\setlength{\extrarowheight}{3pt}
\begin{subtable}[t]{0.99\linewidth}
\centering
\begin{tabular}{|c|c|cccc|cccc|}
\hline
\multirow{2}{*}{\makecell{PDB ID}} & 
\multirow{2}{*}{\makecell{Length\\(nm)}} &
\multicolumn{4}{c|}{\textbf{3D Reconstruction metrics}} & 
\multicolumn{4}{c|}{\textbf{2D metrics}} \\
\cline{3-10}
 & & 
\makecell{CD (nm) \\ $\downarrow$} &
\makecell{HD (nm) \\ $\downarrow$} &
\makecell{F@0.05\% \\ $\uparrow$} &
\makecell{F@0.1\% \\ $\uparrow$} &
\makecell{PSNR \\ $\uparrow$} &
\makecell{SSIM \\ $\uparrow$} &
\makecell{LPIPS \\ $\downarrow$} &
\makecell{MSE \\ $\downarrow$} \\
\hline
6ian & 20.23 & 1.1190 & 5.0263 & 57.24 & 83.07 & 16.09 & 0.5008 & 0.2133 & 0.0260\\
6n1z & 16.97 & 0.8180 & 2.9513 & 60.93 & 87.67 & 18.00 & 0.8222 & 0.1404 & 0.0163\\
6yvu & 31.79 & 0.8488 & 3.9673 & 85.99 & 99.78 & 17.60 & 0.7473 & 0.1358 & 0.0187\\
7uro & 29.03 & 1.0450 & 3.6343 & 71.56 & 95.74 & 17.63 & 0.8607 & 0.1275 & 0.0180\\
6oq5 & 19.99 & 0.8375 & 3.9537 & 67.80 & 93.77 & 16.56 & 0.6711 & 0.1614 & 0.0241\\
\hline
Average & 23.20 & 0.9337 & 3.9066 & 68.30 & 91.61 & 17.18 & 0.7204 & 0.1556 & 0.0206 \\
\hline
\end{tabular}
\caption{}
\label{tab:combined_metrics_virtual_afm}
\end{subtable}

\begin{subtable}[t]{0.99\linewidth}
\centering
\setlength{\extrarowheight}{3pt}
\begin{tabular}{|c|c|cccc|cccc|}
\hline
\multirow{2}{*}{\makecell{Protein \\or PC }} & 
\multirow{2}{*}{\makecell{Length\\(nm)}} &
\multicolumn{4}{c|}{\textbf{3D Reconstruction metrics}} & 
\multicolumn{4}{c|}{\textbf{2D metrics}} \\
\cline{3-10}
 & & 
\makecell{CD (nm) \\ $\downarrow$} &
\makecell{HD (nm) \\ $\downarrow$} &
\makecell{F@0.05\% \\ $\uparrow$} &
\makecell{F@0.1\% \\ $\uparrow$} &
\makecell{PSNR \\ $\uparrow$} &
\makecell{SSIM \\ $\uparrow$} &
\makecell{LPIPS \\ $\downarrow$} &
\makecell{MSE \\ $\downarrow$} \\
\hline
WRC-Rac1 & 19.24 & 1.2890 & 4.0748 & 46.32 & 74.86 & 14.59 & 0.6914 & 0.2991 & 0.0370\\
CasB-E & 16.07 & 0.5995 & 3.1469 & 73.69 & 96.58 & 13.00 & 0.7291 & 0.2570 & 0.0515\\
EIN-HPr & 7.86 & 0.2987 & 1.1861 & 71.95 & 96.49 & 15.18 & 0.7784 & 0.2105 & 0.0324\\
EIN & 7.63 & 0.3320 & 1.2696 & 65.42 & 91.79 & 15.94 & 0.8351 & 0.1653 & 0.0264\\
\hline
Average & 12.70 & 0.6298 & 2.4194 & 64.35 & 89.93 & 14.68 & 0.7585 & 0.2330 & 0.0368 \\
\hline
\end{tabular}
\caption{}
\label{tab:combined_metrics_experimental_afm}
\end{subtable}
\caption{Comparing quantitative performance for proteins and protein complexes. (a) Metrics computed for predictions using virtual AFM images. (b) Metrics computed for predictions using experimental AFM images. Columns are grouped into 3D reconstruction metrics (Chamfer Distance (CD), Hausdorff Distance (HD), and F-scores) and 2D image quality metrics (PSNR, SSIM, LPIPS, MSE). ($\downarrow$) indicates lower value is better, and ($\uparrow$) indicates higher value is better. CD and HD are reported in nanometers. F-scores are computed using thresholds proportional to protein size, where F@0.05 and F@0.1 correspond to 5\% and 10\% of the protein length, respectively. For reference, AFM imaging at a 400nm scan area with 256 pixels yields an effective lateral resolution of $\sim$1.56nm/pixel, indicating that most deviations are within or near experimental resolution.}
\label{tab:combined_metrics}
\end{table*}

\begin{figure}[h!]
    \centering
    \begin{subfigure}[c]{\linewidth}
        \centering
        \includegraphics[width=0.99\linewidth, trim={0.0in 0.25in 0.0in 0.25in}, clip]{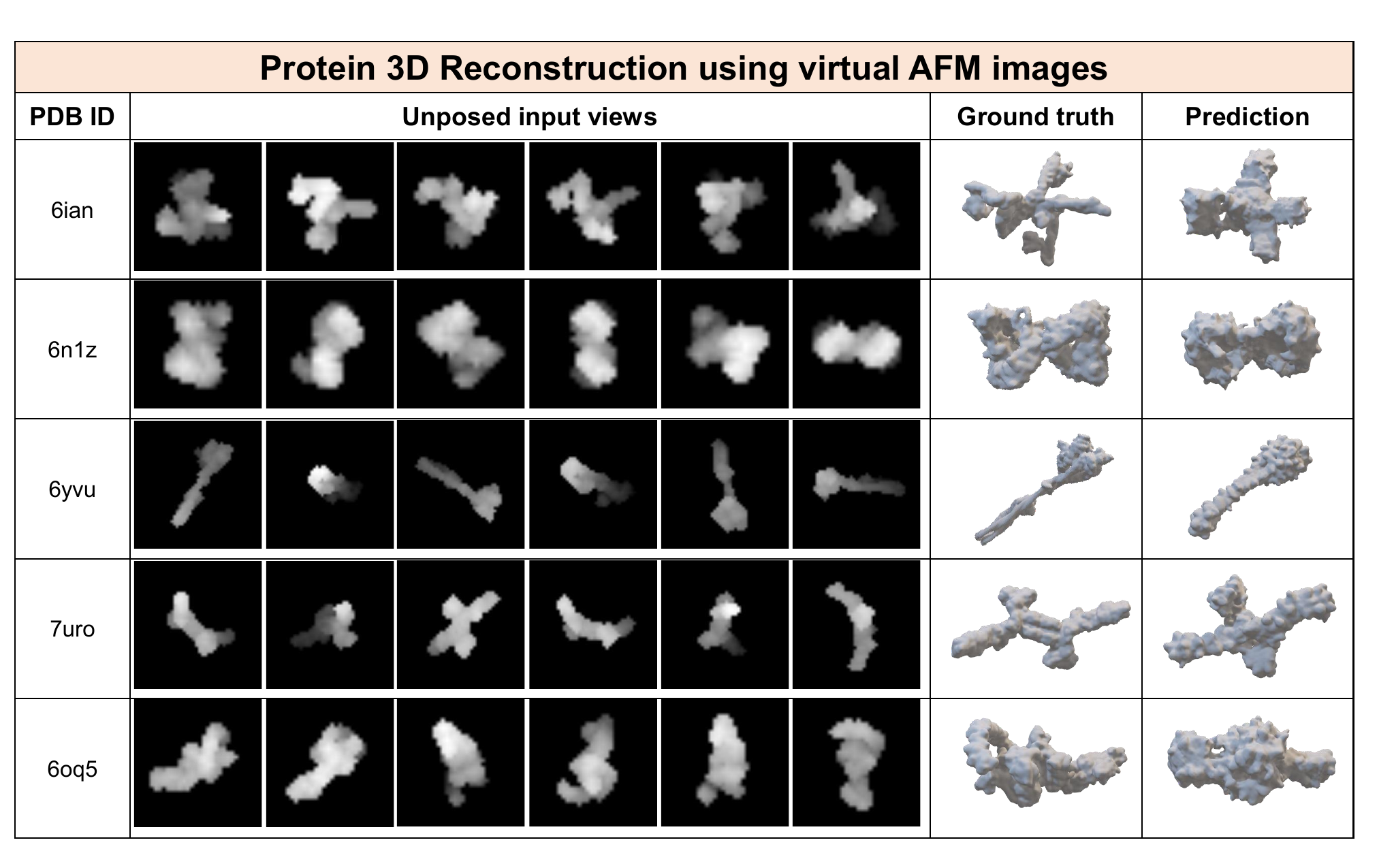}
        \caption{}
        \label{fig:3d_pred_using_virtual_afm}
    \end{subfigure}    
    \begin{subfigure}[c]{\linewidth}
        \centering
        \includegraphics[width=0.99\linewidth, trim={0.0in 0.5in 0.0in 0.5in}, clip]{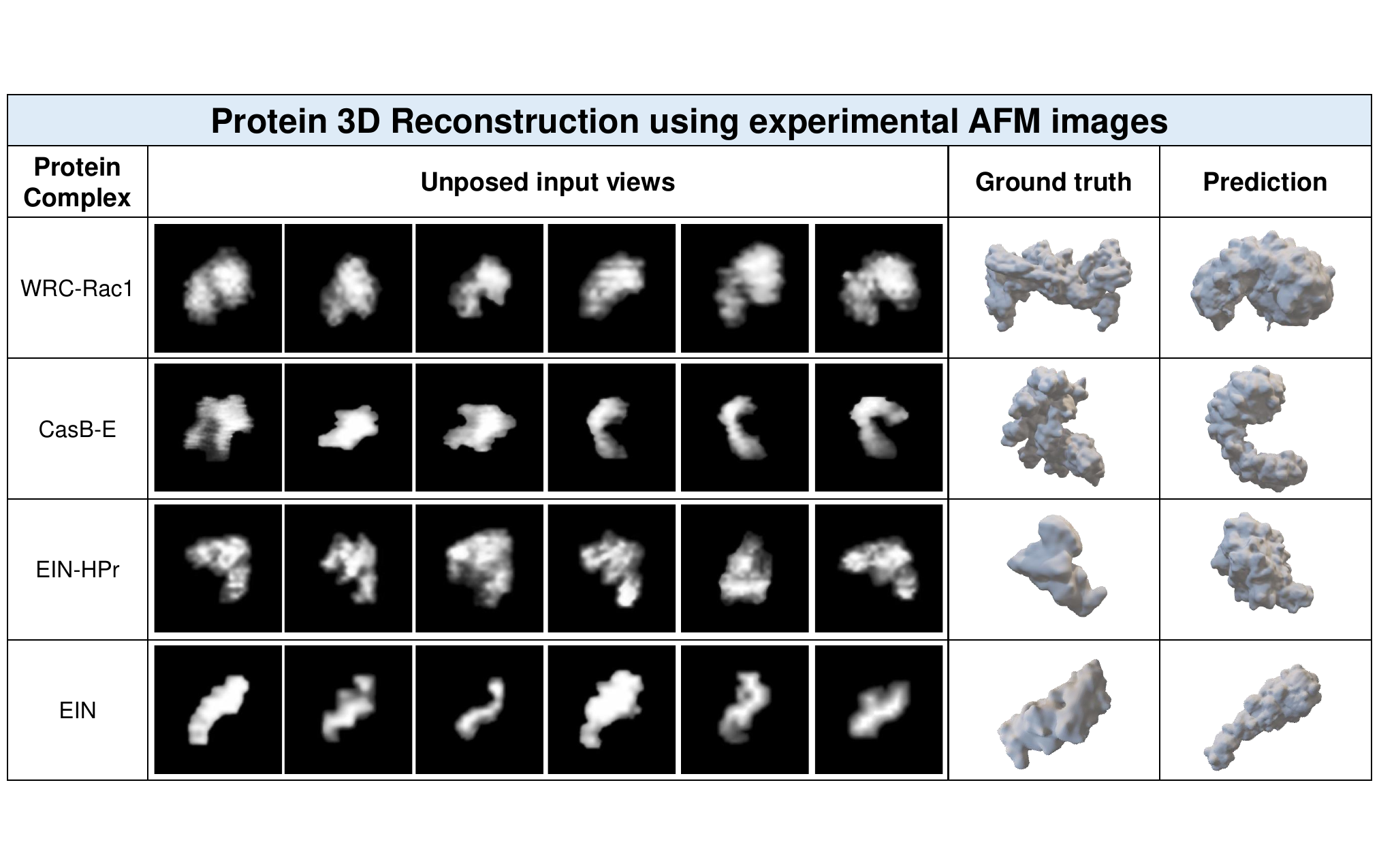}
        \caption{}
        \label{fig:3d_pred_using_experimental_afm}
    \end{subfigure}
    \caption{(a) 3D reconstruction of protein complexes from virtual AFM images. Representative examples of 3D protein prediction using our ProFusion model. For each PDB ID, multiple unposed virtual AFM input views are shown alongside the ground truth 3D structure and the model’s predicted reconstruction.
    (b) Using experimental AFM images to predict Protein complex structures using the ProFusion model on WRC-Rac1, CasB-E, EIN-HPr, and EIN protein and protein complex structures.}
    \label{fig:combined_3d_preds}
\end{figure}

\begin{figure}[ht!]
    \centering
    \begin{subfigure}[t]{\linewidth}
        \centering
        \includegraphics[width=0.94\linewidth, trim={0.15in 0.0in 0.15in 0.0in}, clip]{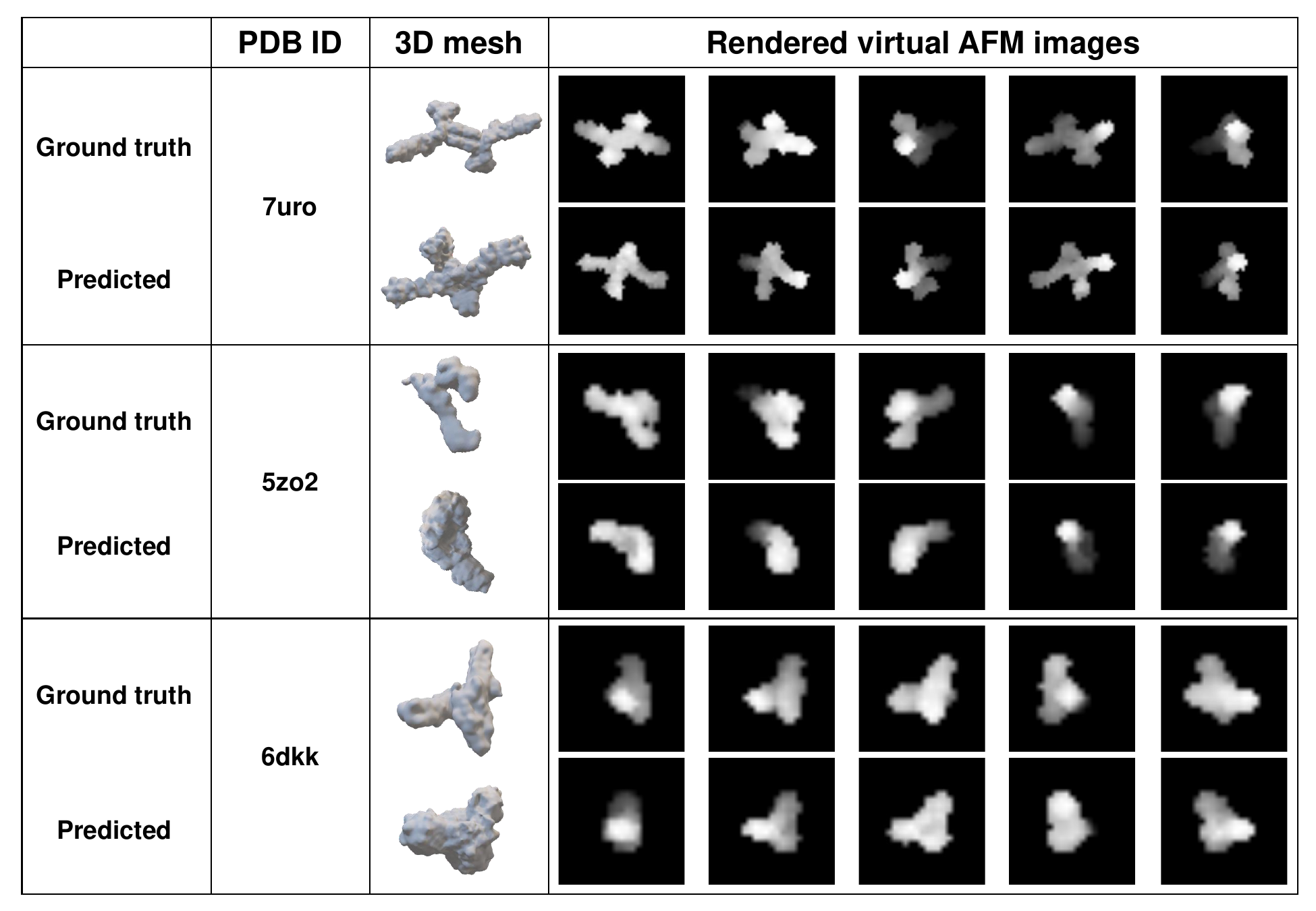}
        \caption{}
        \label{fig:vafm_renderings_multimers}
    \end{subfigure}
    \begin{subfigure}[t]{\linewidth}
        \centering
        \includegraphics[width=0.94\linewidth, trim={0.15in 1.25in 0.15in 1.25in}, clip]{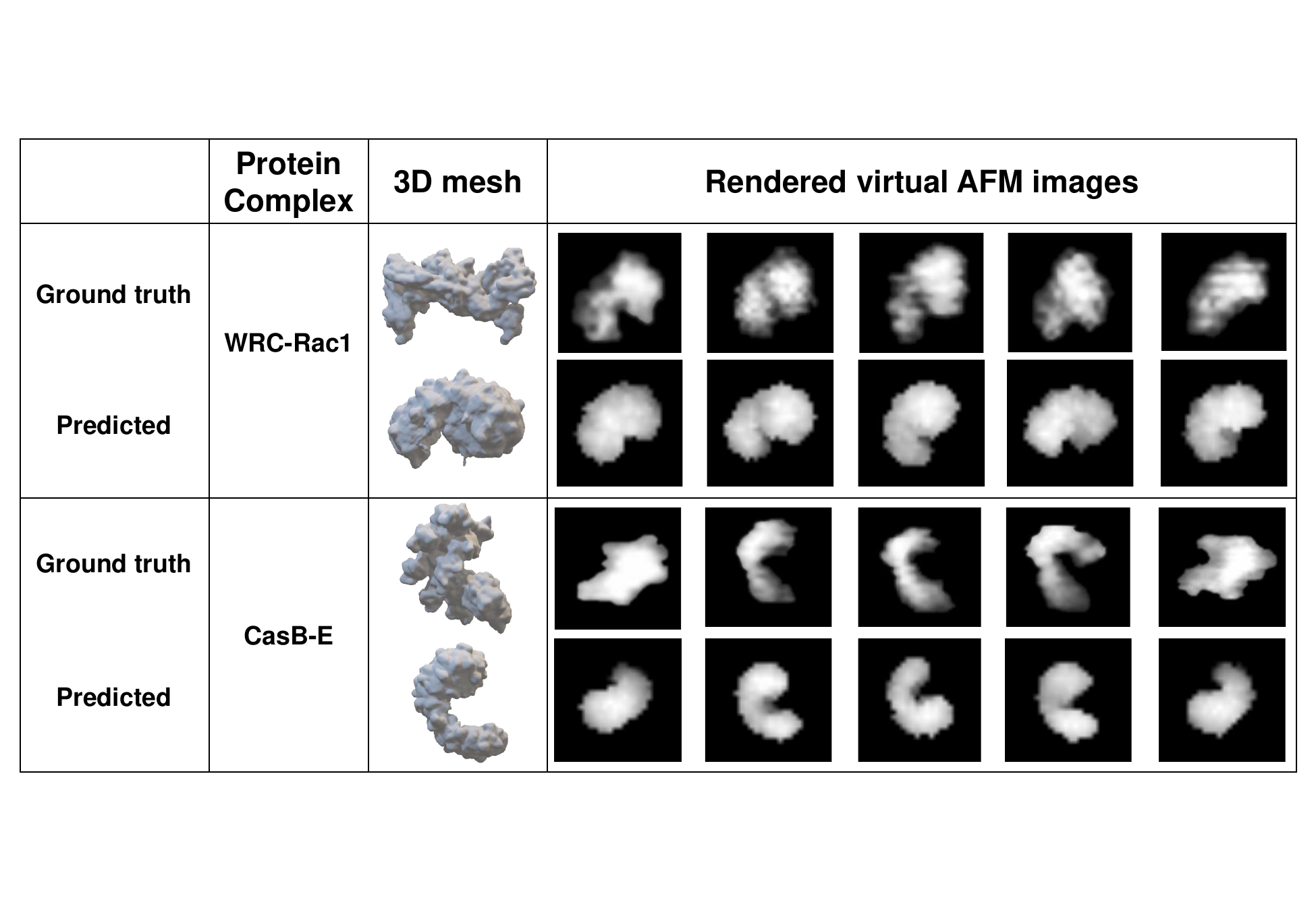}
        \caption{}
        \label{fig:vafm_renderings_experimental_afm}
    \end{subfigure}
    \caption{Comparison of virtual AFM renderings of predicted and ground-truth protein structures. 
    (a) 3D protein structures are first reconstructed from virtual AFM images, after which 2D virtual AFM renderings of the reconstructed structures are generated using our \textit{virtual} AFM tool and compared with the corresponding ground-truth 2D images.
    (b) 3D reconstructions of the WRC–Rac1 and CasB–E protein complexes are obtained from experimental AFM images, and their virtual AFM renderings are compared against the corresponding experimental AFM views.
    These visualizations highlight the model’s ability to produce realistic reconstructions consistent with experimental AFM images.}
    \label{fig:vafm_combined}
\end{figure}

We train our DL model using a large-scale dataset of virtual AFM images of proteins and PCs. The details of the DL model architecture and the dataset generation using the \textit{virtual} AFM framework are presented in the \secref{sec:profusion} and \secref{sec:virtualAFM}, respectively. Model performance is then evaluated on a test set, which is not used during training, to ensure unbiased assessment. We evaluate 3D structure predictions using both virtual and experimental AFM images to demonstrate that our method generalizes effectively to experimental AFM images despite being trained solely on virtual images. Ultimately, our primary objective is to assess how accurately the model can reconstruct protein 3D structures from experimental AFM images. Evaluation focuses on two aspects: (1) 3D reconstruction accuracy of the predicted structures and (2) visual and perceptual quality of the rendered AFM images from these reconstructions.

\subsubsection*{3D Reconstruction Evaluation}

We evaluate our approach using quantitative and qualitative analyses of the reconstructed 3D structures. For quantitative evaluation, we first export the predicted 3D meshes from the NeRF representation and align them with the ground-truth meshes using a rigid transformation to account for any rotational or translational offsets. Predicted and ground-truth meshes are normalized to fit within a unit cube centered at the origin to ensure scale invariance. We then uniformly sample 10,000 points from each mesh surface, which provides a consistent point-based representation for geometric comparison.

We compute three widely used metrics for 3D structure assessment: Chamfer Distance (CD), Hausdorff Distance (HD), and the F-score~\citep{arno2017F-score}. CD measures the average bidirectional nearest-neighbor distance between the predicted and ground-truth point sets, capturing overall reconstruction accuracy. HD quantifies the maximum minimum distance between the two point sets, highlighting worst-case deviations and sensitivity to local errors. F-score evaluates the harmonic mean of precision and recall under a specified distance threshold, effectively balancing the accuracy and completeness of the reconstructed surfaces. In this context, precision measures the reconstruction accuracy as the fraction of predicted points that lie close to the ground-truth points. At the same time, recall quantifies the reconstruction completeness as the fraction of ground-truth points that are covered by the prediction. Together with CD and HD, the F-score offers a comprehensive assessment of reconstruction quality.

As summarized in \tabref{tab:combined_metrics_virtual_afm}, our model achieves an average CD of 0.93 nm and HD of 3.91 nm, which are within or close to the AFM imaging resolution (for example, 1.95 nm/pixel for a 500 nm scan with 256 pixels). The F-scores reach 68.3\% at 5\% and 91.6\% at 10\% of the protein length, confirming high geometric fidelity and surface completeness. Most deviations lie near or below the resolution limit of AFM, indicating that reconstructions are experimentally meaningful.

To assess generalization to experimental data, we apply the model (trained only on virtual AFM images) to experimental AFM images of proteins and PCs. \tabref{tab:combined_metrics_experimental_afm} reports the quantitative results. Despite a domain shift, the model achieves an average CD of 0.63 nm and HD of 2.42 nm, demonstrating that the predicted 3D structures remain highly accurate within the resolution limits of AFM experimental imaging (as discussed in \secref{sec:afm_scan_params}). The F-scores are 64.4\% at 5\% and 89.9\% at 10\%, only slightly lower than the performance on virtual AFM inputs, showing that the model generalizes well to experimental AFM images.

\clearpage

To qualitatively evaluate our method, we visualize the reconstructed 3D structures of the proteins and PCs and compare them to their ground-truth structures. \figref{fig:3d_pred_using_virtual_afm} shows reconstructions obtained from virtual AFM images, while \figref{fig:3d_pred_using_experimental_afm} presents results from experimental AFM images. In both cases, the predicted meshes accurately capture the global surface topology of the protein structures, closely resembling the corresponding ground-truth structures. This high level of structural similarity demonstrates that our method can reliably reconstruct protein shapes from both virtual and experimental AFM images, even without explicit pose information to guide the 3D reconstruction.

\subsubsection*{2D Virtual AFM Evaluation}

Furthermore, we generate virtual AFM images for the predicted 3D structures and compare them with the AFM images rendered for the ground-truth structures. Virtual AFM images are generated from multiple viewing directions and evaluated qualitatively and quantitatively. For selected PCs such as WRC-Rac1, CasB-E, EIN-HPr, and EIN, we directly compare the generated virtual AFM images with their corresponding experimental AFM images. For quantitative evaluation, we use standard image reconstruction metrics, including Peak Signal-to-Noise Ratio (PSNR), Structural Similarity Index (SSIM)~\citep{Wang2004SSIM}, Mean Square Error (MSE), and Learned Perceptual Image Patch Similarity (LPIPS)~\citep{Zhang2018LPIPS}. 

\tabref{tab:combined_metrics_virtual_afm} reports results for five proteins reconstructed from virtual AFM inputs. On average, our method achieves 17.18dB PSNR and 0.72 score for SSIM, indicating high structural fidelity between the predicted and ground-truth virtual AFM images. The low LPIPS score of 0.16 and MSE of 0.021 confirm that deviations are minimal and largely imperceptible at typical AFM resolutions.

\tabref{tab:combined_metrics_experimental_afm} summarizes the comparison metrics between the rendered virtual AFM projections of our predicted 3D structures with the experimental AFM images. On average, the PSNR value of 14.68dB and an SSIM of 0.76 indicate that the reconstructed structures produce 2D views that remain structurally consistent with experimental AFM images. As expected, the PSNR is slightly lower due to inherent experimental noise, tip convolution effects, and minor misalignments in experimental AFM imaging. Nevertheless, SSIM remains high, and the LPIPS score of 0.23 and MSE value of 0.036 confirm that the rendered virtual AFM images retain substantial perceptual similarity to the experimental AFM images.

To complement the quantitative results, \figref{fig:vafm_renderings_multimers} compares virtual AFM images generated from reconstructed protein structures with those from the ground truth. Specifically, we first use ProFusion to reconstruct 3D protein structures from virtual AFM image inputs. Then, we apply our \textit{virtual} AFM tool to render 2D images of the reconstructed structures, which are directly compared against the ground-truth virtual AFM images. For the WRC–Rac1 and CasB–E protein complexes, we extend this evaluation to experimental AFM images: 3D structures are reconstructed from experimental AFM images, and their virtual AFM renderings are compared with the corresponding experimental AFM images, as shown in \figref{fig:vafm_renderings_experimental_afm}.

Overall, this 2D evaluation reinforces the 3D reconstruction results, showing that our approach not only predicts topologically accurate 3D protein structures but also produces virtual AFM images that closely match both virtual and experimental AFM images corresponding to ground truth 3D structures. This consistency highlights the practical application of our method for predicting the 3D structures of PCs directly from experimental AFM images.

\section{Discussion}\label{discussion}

In this work, we propose the ProFusion, a novel DL-based framework for predicting the 3D structures of proteins and protein complexes using AFM images. Our approach focuses on training the neural network on unposed, multi-view AFM images to synthesize novel views conditioned on arbitrary query poses. These synthesized views and their corresponding pose information are then used to train an instance-specific NeRF model, enabling accurate 3D reconstruction of the protein structure. The AFM experimental observations underscore the value of multi-view topographical imaging as a complementary tool for protein structure analysis. Even without knowing the orientation of each molecule \textit{a priori}, the collection of random-view AFM images captures the expected structural features of the protein complexes. Importantly, having multiple orientations for each protein due to random attachment on the mica surface during sample preparation is crucial to understanding the overall 3D structure than any single view alone. This multi-view capability of AFM highlights its unique strength in structural characterization. 

The height and roughness measurements from AFM also contribute valuable insights about the protein shape. Height variation can indicate protein–protein interaction and complex stability in solution, while nanoscale roughness affects protein adsorption onto surfaces. These factors influence binding strength, distribution of contact sites, and structural alignment. For example, the EIN-HPr complex shows a slight increase and a clear elongation compared to EIN alone, consistent with structural reorganization upon binding. This interaction, centered around key residues (His189 of EIN and His15 of HPr), supports previous structural models and highlights the value of AFM in probing dynamic protein interfaces reported by \citet{schwieters2010solution}. Such features are important for applications requiring stable protein immobilization, especially for our application of 3D reconstruction~\citep{rechendorff2006enhancement, scopelliti2010effect}.

Finally, we developed a \textit{virtual} AFM simulation tool that emulates the physical AFM imaging process through a rasterization-based rendering pipeline to support the training of DL models. This tool generates realistic, multi-view AFM-like images from protein structures in PDB format. Using this tool, we created a large-scale dataset comprising approximately 542,000 proteins, each represented by multiple \textit{virtual} AFM views. The \textit{virtual} AFM tool not only facilitates the training of DL models on proteins but also serves as a valuable resource for generating synthetic AFM data for a wide range of materials, enabling further research in AFM-guided structural analysis. We will release the dataset along with this paper. We utilize the synthetically generated multi-view AFM dataset to train the ProFusion network for novel view synthesis and 3D structure reconstruction tasks. To assess the effectiveness and generalizability of our approach, we conduct extensive quantitative and qualitative validation on both virtual and experimental AFM images across various proteins and protein complexes. This includes evaluating the ability of the model to reconstruct 3D structures from unposed AFM images, demonstrating that our method can effectively learn structural representations even without explicit pose information. Our results highlight the potential of combining AFM data and DL to advance protein structure prediction in scenarios where experimental data is limited or costly to acquire.

\section*{Acknowledgments}
This work used the Delta system at the National Center for Supercomputing Applications through allocation CIS240344 from the Advanced Cyberinfrastructure Coordination Ecosystem: Services and Support (ACCESS) program, which is supported by National Science Foundation (NSF) grants 2138259, 2138286, 2138307, 2137603, and 2138296. This work was also partly supported by the National Institutes of Health R35 GM128786 to B.C.

\section*{Author Contributions}

Conceptualization: A.S., A.K., S.S.; Methodology: J.R., S.S., A.K.; Software (model training and Virtual AFM): J.R.; Result analysis: J.R.; Data collection (AFM experiments): H.H., A.S.; Resources (protein and protein complex samples): M.O., B.F., S.Y., D.S., V.V., B.C.; Writing – original draft: J.R., H.H.; Writing – review and editing: J.R., H.H., S.S., A.K., A.S.; Supervision: A.S., A.K.; Approval of manuscript: All authors.

\section*{Data Availability}
The datasets generated during and/or analyzed during the current study will be made public upon acceptance of the paper. 

\section*{Code Availability}
The source code for this study is available via GitHub at  \url{https://github.com/idealab-isu/ProFusion} and \url{https://github.com/idealab-isu/VirtualAFM}.

\section*{Competing interests}
The authors declare no competing interests.

\pagebreak

\bibliographystyle{elsarticle-harv}
\bibliography{references}

\appendix
\newpage
\setcounter{page}{1}
\renewcommand\thefigure{S.\arabic{figure}}    
\renewcommand\thesection{\Alph{section}}    
\renewcommand\thetable{S.\arabic{table}}    
\setcounter{figure}{0}
\setcounter{table}{0}
\setcounter{section}{0}

\begin{center}
{\usefont{OT1}{phv}{b}{n}\selectfont\Large{ProFusion: 3D Reconstruction of Protein Complex Structures from Multi-view AFM Images

Supplementary Material
}}
\end{center}

\section{Protein Purification}\label{sec:protein_purification}

\textbf{CasB-E purification:} CasB-D complex is expressed using pSV272-His6-tag and CasE is expressed using pCDF-1b-no-tag construct, crRNA is provided using pACYC low copy plasmid and pRSF high copy plasmid. All plasmids are transformed into BL21(DE3) cells simultaneously. Cultures are grown to 0.5 OD600 at 37°C and induced overnight at 20°C with 0.5 mM isopropyl $\beta$-D-1 thiogalactopyranoside (IPTG). Cells are lysed in the lysis buffer (50 mM Sodium phosphate dibasic, 500 mM Sodium chloride, 5\% glycerol, 10 mM imidazole, pH 8.0, cOmplete™, EDTA-free Protease Inhibitor Cocktail, 1 mM DTT) using a homogenizer. Cell debris is removed by centrifuging the cell lysate at 19,000 rpm for 30 minutes. Supernatant is collected and run through HisPur Ni-NTA affinity resin in recommended buffers (Thermo Fisher Scientific) to purify CasB-E. The purified CasB-E is cleaved by tobacco etch virus (TEV) protease overnight at 4ºC to remove the His6-tag in dialysis buffer (50 mM Sodium phosphate dibasic, 500 mM Sodium chloride, 5\% glycerol, 2 mM DTT). The cleaved CasB-E is flowed through a Ni-NTA column to remove uncleaved protein, concentrated to 1 mL, and purified on a Superdex 200 column in a buffer containing 20 mM Tris (pH 7.5), 100 mM NaCl, 5\% glycerol, and 1 mM DTT.

\section{Surface Roughness Analysis Using AFM}

\begin{figure}[ht!]
    \centering
   \includegraphics[width=0.7\linewidth, trim={0.0in 0.0in 0.0in 0.0in},clip]{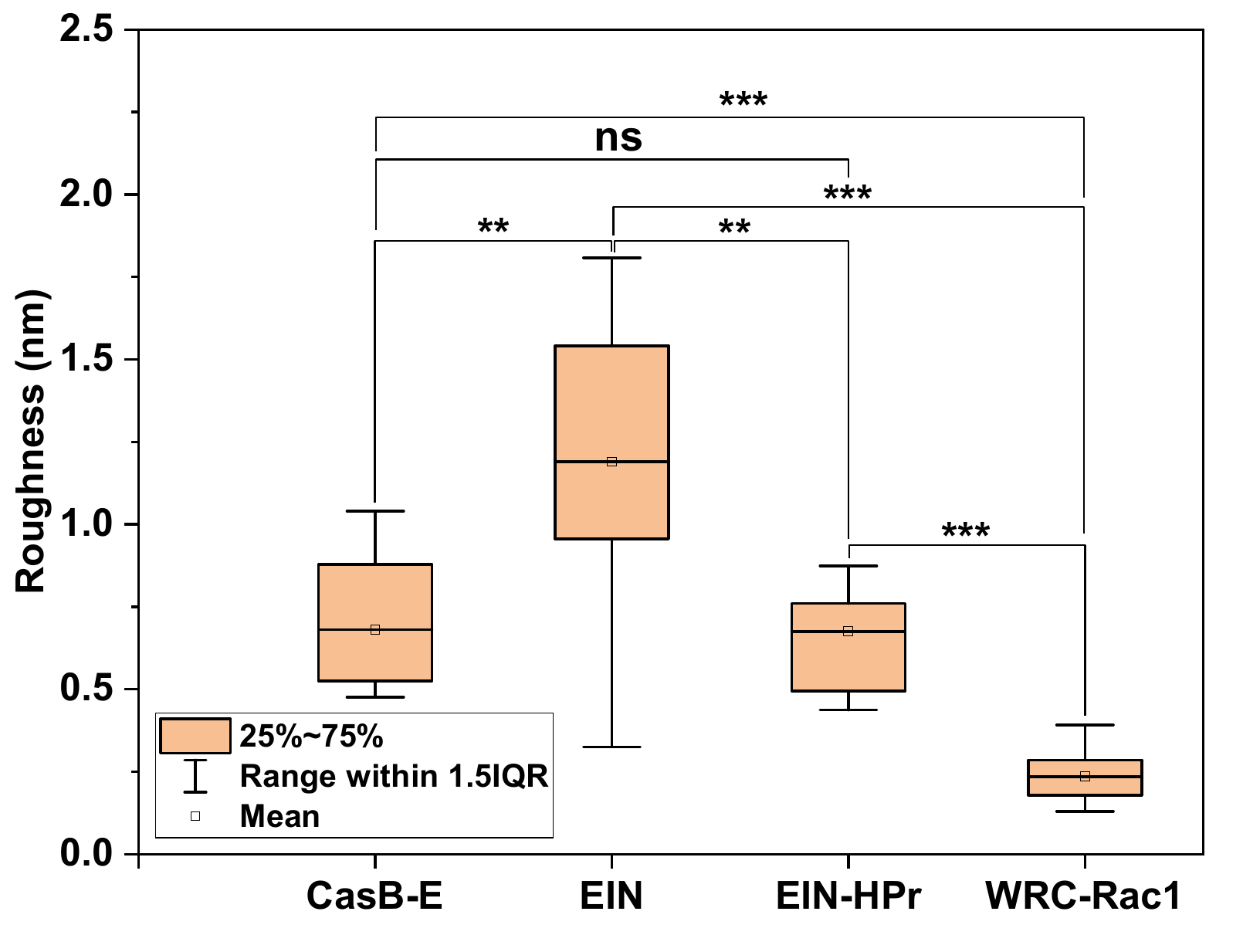}
    \caption{Quantitative analysis of surface roughness. EIN exhibits the highest value (1.19 nm), whereas WRC-Rac1 shows the lowest (0.24 nm). All measurements are obtained using the NanoScope Analysis software 3.00.Statistical significance is denoted as follows: *~\textit{p}~$<$~0.05,\quad **~\textit{p}~$\leq$~0.01,\quad ***~\textit{p}~$<$~0.0001,\quad \textit{p}~$\geq$~0.05 = not significant (ns).}
    \label{fig:Roughness R1}
\end{figure}

Surface roughness analysis (\figref{fig:Roughness R1}) demonstrates clear differences among the protein complexes. EIN shows the highest roughness value (1.19 nm), reflecting greater surface irregularity, whereas WRC-Rac1 exhibits the lowest value (0.24 nm), consistent with its relatively smooth topography. CasB-E and EIN-HPr display intermediate roughness profiles. These quantitative variations highlight distinct nanoscale surface properties across the complexes, which may influence protein stability and substrate interaction.

\section{AFM Schematic}

\begin{figure}[ht!]
    \centering
   \includegraphics[width=0.5\linewidth, trim={0.0in 0.0in 0.0in 0.0in},clip]{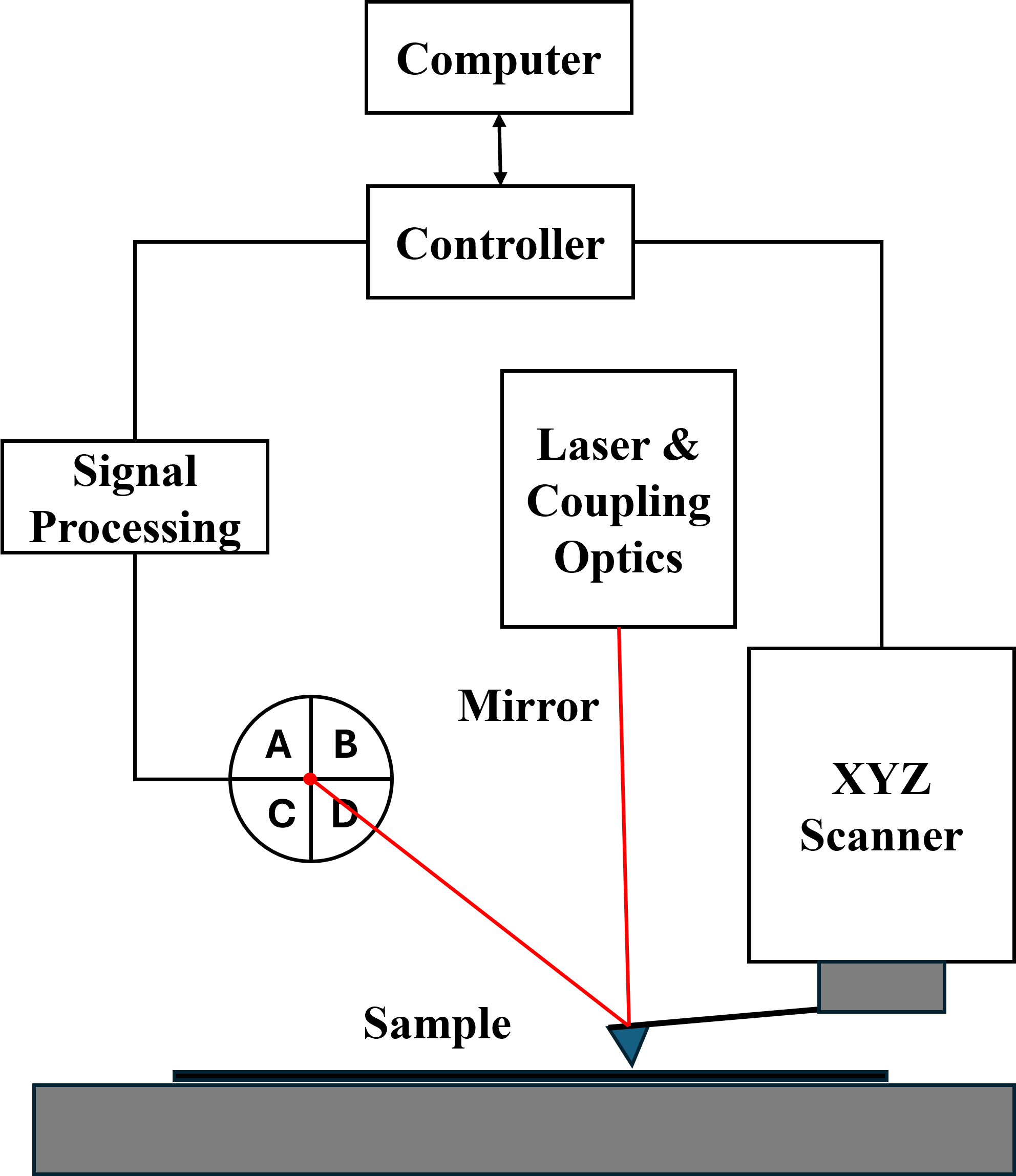}
    \caption{Schematic depiction of an atomic force microscope (not to scale): A laser from a diode is aimed at the tip of the cantilever. The reflected light is captured by a quadrant photo-diode, which measures the vertical and lateral movements of the cantilever. The system's controller uses a feedback loop to adjust the cantilever’s position depending on the set-point chosen by the user. Data is collected and analyzed by a computer to configure the controller settings and generate images.}
    \label{fig:AFM setup}
\end{figure}

\figref{fig:AFM setup} illustrates the working principle of an atomic force microscope (AFM). A laser beam is directed onto the back of the cantilever and reflected to a quadrant photo-diode, where vertical and lateral deflections are detected. The feedback controller continuously adjusts the cantilever’s position through the XYZ scanner to maintain the set-point force. The computer then analyzes the processed signal to generate high-resolution images of the sample surface.

\section{Ablation Study: Varying Number of Input Views}

\begin{table}[h!]
\centering
\begin{tabular}{|c|c|c|c|c|c|c|c|c|}
  \hline
  $\#$ views & \makecell{CD (nm) \\ ($\downarrow$)} & \makecell{HD (nm) \\ ($\downarrow$)} & \makecell{F-score@0.05\% \\ ($\uparrow$)}& \makecell{F-score@0.1\% \\ ($\uparrow$)} & \makecell{PSNR \\ ($\uparrow$)} & \makecell{SSIM \\ ($\uparrow$)} & \makecell{LPIPS \\ ($\downarrow$)} & \makecell{MSE \\ ($\downarrow$)} \\
  \hline
  
  1 & 0.9772 & 4.3250 & 60.42 & 83.98 & 16.52 & 0.7365 & 0.1646 & 0.0243\\
  \hline
  3 & 0.9728 & 3.9857 & 59.34 & 84.27 & 17.04 & 0.6944 & 0.1640 & 0.0216 \\
  \hline
  6 & 0.9170 & 3.6189 & 61.50 & 86.26 & 17.36 & 0.7515 & 0.1551 & 0.0198 \\
  \hline
\end{tabular}
\caption{Comparing quantitative performance for proteins and protein complexes using different numbers of input views. (a) Metrics computed for predictions using virtual AFM images. (b) Metrics computed for predictions using experimental AFM images. Columns are grouped into 3D reconstruction metrics (Chamfer Distance (CD), Hausdorff Distance (HD), and F-scores) and 2D image quality metrics (PSNR, SSIM, LPIPS, MSE). ($\downarrow$) indicates lower value is better, and ($\uparrow$) indicates higher value is better. CD and HD are reported in nanometers. F-scores are computed using thresholds proportional to protein size, where F@0.05 and F@0.1 correspond to 5\% and 10\% of the protein length, respectively.}
\label{tab:combined_metric_diff_num_views}
\end{table}

\begin{figure}[ht!]
  \centering
  \includegraphics[width=0.9\linewidth, trim={0.5in 0.0in 0.0in 0.25in},clip]{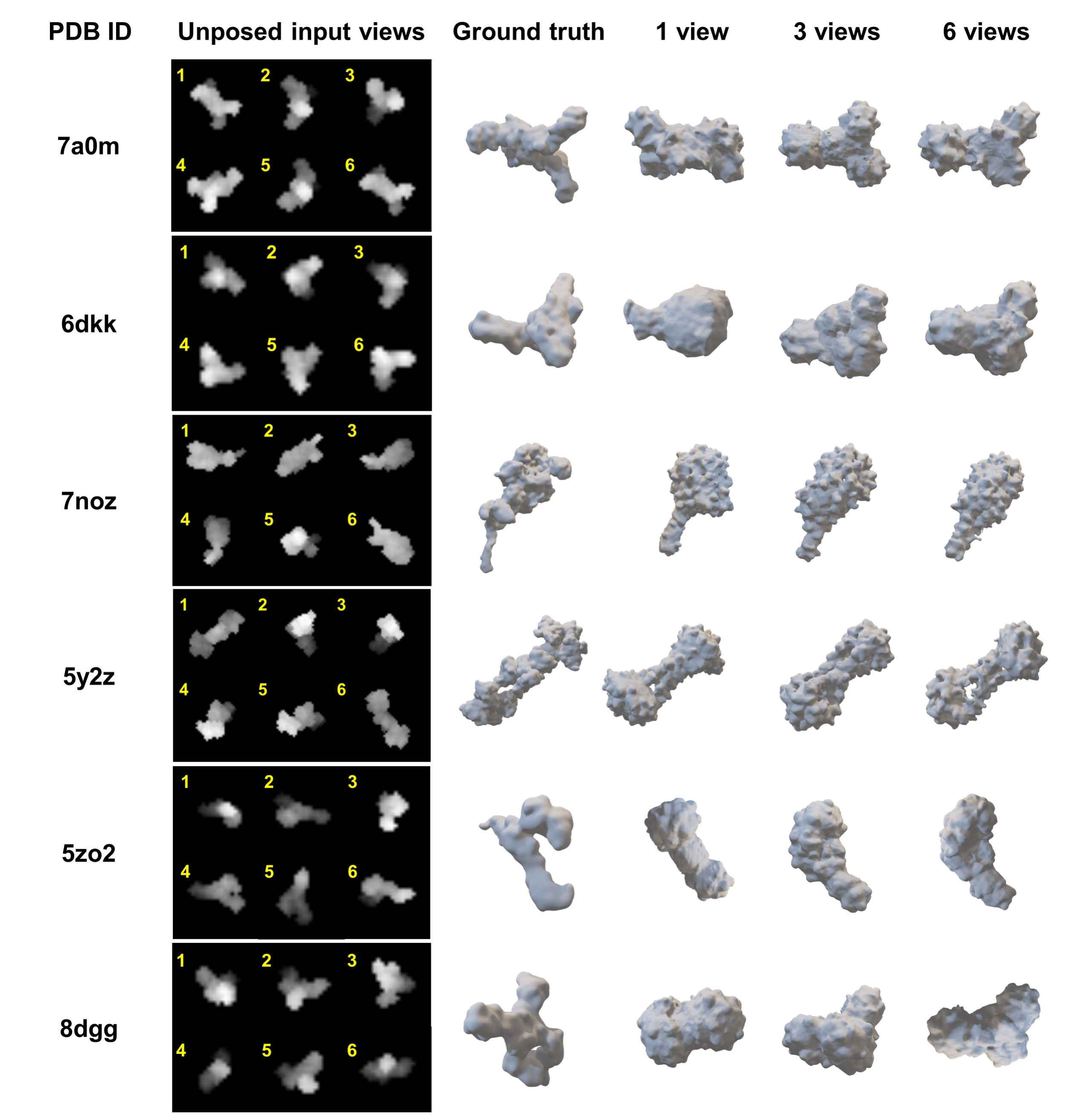}
   \caption{Comparing the 3D predicted structures of PCs corresponding to different numbers of input views.}
   \label{fig:diff_num_views_extra}
\end{figure}

We conduct an ablation study to assess how the number of input views impacts 3D protein structure reconstruction. Inference is performed using one, three, and six input views, and results are summarized in \tabref{tab:combined_metric_diff_num_views}. For 3D metrics, we observe clear gains as the number of views increases. CD improves by approximately $6\%$, and HD improves by $16\%$ when moving from 1 to 6 input views. F-score values also increase consistently at both threshold levels, indicating better geometric accuracy and surface completeness. For 2D metrics, a similar trend is seen in the quality of rendered virtual AFM images. PSNR rises from 16.52 dB (1 view) to 17.36 dB (6 views), MSE decreases from 0.0243 to 0.0198, and LPIPS improves from 0.1646 to 0.1551, reflecting better perceptual alignment with the ground truth. SSIM also increases overall, reaching 0.75 with six views, highlighting improved structural fidelity. These results confirm that incorporating more views leads to higher geometric accuracy, structural consistency, and perceptual realism in the reconstructed protein structures. To illustrate these improvements, \figref{fig:diff_num_views_extra} presents qualitative comparisons of predicted and ground-truth structures across different numbers of input views.

\end{document}